\documentclass[format=acmsmall, review=false, screen=true, authorversion=true]{acmart}

\usepackage{booktabs} 

\usepackage[ruled]{algorithm2e} 

\SetAlFnt{\small}
\SetAlCapFnt{\small}
\SetAlCapNameFnt{\small}
\SetAlCapHSkip{0pt}
\IncMargin{-\parindent}
\def\etal{\emph{et al.}~}
\usepackage{multirow}

\acmJournal{TOMM}
\acmVolume{14}
\acmNumber{2}
\acmArticle{48}
\acmYear{2018}
\acmMonth{5}
\copyrightyear{2018}

\setcopyright{acmlicensed}
\usepackage{url}
\usepackage{graphicx}
\usepackage{amsmath}
\usepackage{amssymb}
\usepackage[caption=false]{subfig}
\usepackage{hyperref}
\usepackage{color, colortbl}
\usepackage{mathtools}
\usepackage{array}
\usepackage[export]{adjustbox}

\acmDOI{3210458.3177745}

\received{November 2017}
\received[accepted]{May 2018}

\begin{document}
\title{Paying More Attention to Saliency: Image Captioning with Saliency and Context Attention}  
\author{Marcella Cornia}
\affiliation{%
  \institution{University of Modena and Reggio Emilia}
  \streetaddress{Via P. Vivarelli, 10}
  \city{Modena}
  \state{}
  \postcode{41125}
  \country{Italy}}
\author{Lorenzo Baraldi}
\affiliation{%
  \institution{University of Modena and Reggio Emilia}
  \streetaddress{Via P. Vivarelli, 10}
  \city{Modena}
  \state{}
  \postcode{41125}
  \country{Italy}}
\author{Giuseppe Serra}
\affiliation{%
  \institution{University of Udine}
  \streetaddress{}
  \city{Udine}
  \state{}
  \postcode{}
  \country{Italy}}
\author{Rita Cucchiara}
\affiliation{%
  \institution{University of Modena and Reggio Emilia}
  \streetaddress{Via P. Vivarelli, 10}
  \city{Modena}
  \state{}
  \postcode{41125}
  \country{Italy}}

\begin{abstract}
Image captioning has been recently gaining a lot of attention thanks to the impressive achievements shown by deep captioning architectures, which combine Convolutional Neural Networks to extract image representations, and Recurrent Neural Networks to generate the corresponding captions. At the same time, a significant research effort has been dedicated to the development of saliency prediction models, which can predict human eye fixations. Even though saliency information could be useful to condition an image captioning architecture, by providing an indication of what is salient and what is not, research is still struggling to incorporate these two techniques. In this work, we propose an image captioning approach in which a generative recurrent neural network can focus on different parts of the input image during the generation of the caption, by exploiting the conditioning given by a saliency prediction model on which parts of the image are salient and which are contextual. We show, through extensive quantitative and qualitative experiments on large scale datasets, that our model achieves superior performances with respect to captioning baselines with and without saliency, and to different state of the art approaches combining saliency and captioning.
\end{abstract}




\thanks{This work is partially supported by ``Citt\`a educante'' (CTN01\_00034\_393801) of the National Technological Cluster on Smart Communities (cofunded by the Italian Ministry of Education, University and Research - MIUR) and by the project ``JUMP - Una piattaforma sensoristica avanzata per rinnovare la pratica e la fruizione dello sport, del benessere, della riabilitazione e del gioco educativo'', funded by the Emilia-Romagna region within the POR-FESR 2014-2020 program. 
We acknowledge the CINECA award under the ISCRA initiative, for the availability of high performance computing resources and support. We also gratefully acknowledge the support of NVIDIA Corporation with the donation of the GPUs used for this research.

  Author's addresses: M. Cornia {and} L. Baraldi {and} R. Cucchiara, Department of Engineering ``Enzo Ferrari'', University of Modena and Reggio Emilia, Modena, Italy; G. Serra is with the Department of Computer Science, Mathematics and Physics, University of Udine, Udine, Italy.}

\maketitle

\renewcommand{\shortauthors}{M. Cornia, L. Baraldi, G. Serra, and R. Cucchiara}


\section{Introduction}
A core problem in computer vision and artificial intelligence is that of building a system that can replicate the human ability of understanding a visual stimuli and describing it in natural language. Indeed, this kind of system would have a great impact on society, opening up to a new progress in human-machine interaction and collaboration.
Recent advancements in computer vision and machine translation, together with the availability of large datasets, have made it possible to generate natural sentences describing images.  
In particular, deep image captioning architectures have shown impressive results in discovering the mapping between visual descriptors and words~\cite{karpathy2015deep,vinyals2015show,icml2015xuc15,you2016image}. They combine Convolutional Neural Networks (CNNs), to extract an image representation, and Recurrent Neural Networks (RNNs), to build the corresponding sentence. 

While the progress of these techniques is encouraging, the human ability in the construction and formulation of a sentence is still far from being adequately emulated in today's image captioning systems. When humans describe a scene, they look at an object before naming it in a sentence~\cite{griffin2000eyes}, and they do not focus on each region with the same intensity, as selective mechanisms attract their gaze on saliency and relevant parts of the scene~\cite{Rensink2000}. Also, they care about the context using peripheral vision, so that the description of an image alludes not only to the main objects in the scene, and to how they relate to each other, but also to the context in which they are placed in the image. 

An intensive research effort has been carried out in the computer vision community to predict where humans look in an image. This task, called saliency prediction, has been tacked in early works by defining hand-crafted features that capture low-level cues such as color and texture or higher-level concepts such as faces, people and text~\cite{itti1998model,borji2012boosting,judd2009learning}. Recently, with the advent of deep neural networks and large annotated datasets, saliency prediction techniques have obtained impressive results generating maps that are very close to the ones computed with eye-tracking devices~\cite{mlnet2016,jetley2016end,huang2015salicon}.

Despite the encouraging progress in image captioning and visual saliency, and their close connections, these two fields of research have remained almost separate. In fact, only few attempts have been recently presented in this direction~\cite{sugano2016seeing,tavakoli2017can}.
In particular, Sugano~\etal~\cite{sugano2016seeing} presented a gaze-assisted attention mechanism for image caption based on human eye fixations (\textit{i.e.} the static states of gaze upon a specific location). Although this strategy confirms the importance of using eye fixations, it requires gaze information from a human operator. Therefore, it can not be applied on general visual data archives, in which this information is missing. 
To overcome this limit, Tavakoli~\etal~\cite{tavakoli2017can} presented an image captioning method based on saliency maps, which can be automatically predicted from the input image.

In this paper we present an approach which incorporates saliency prediction to effectively enhance the quality of image description. We propose a generative Recurrent Neural Network architecture which can focus on different regions of the input image by means of an attentive mechanism. This attentive behaviour, differently from previous works~\cite{icml2015xuc15}, is conditioned by two different attention paths: the former focused on salient spatial regions, predicted by a saliency model, and the latter focused on contextual regions, which are computed as well from saliency maps. Experimental results on five public image captioning datasets (SALICON, COCO, Flickr8k, Flickr30k and PASCAL-50S), demonstrate that our solution is able to properly exploit saliency cues. Also, we show that this is done without losing the key properties of the generated captions, such as their diversity and the vocabulary size. By visualizing the states of both attentive paths, we finally show that the trained model has learned to attend to both salient and contextual regions during the generation of the caption, and that attention focuses produced by the network effectively correspond, step by step, to generated words.

To sum up, our contributions are as follows. First, we show that saliency can enhance image description, as it provides an indication of what is salient and what is context. Second, we propose a model in which the classic machine attention approach is extended to incorporate two attentive paths, one for salient regions and one for context. These two paths cooperate together during the generation of the caption, and show to generate better captions according to automatic metrics, without loss of diversity and size of the dictionary. Third, we qualitatively show that the trained model has learned to attend to both salient and contextual regions in an appropriate way.

\section{Related work}
In this section, we review the literature related to saliency prediction and image captioning. We also report some recent works which investigate the contribution of saliency for generating natural language descriptions.

\subsection{Visual saliency prediction}
Saliency prediction has been extensively studied by the computer vision community and, in the last few years, has achieved a considerable improvement thanks to the large spread of deep neural networks~\cite{kummerer2014deep,huang2015salicon,kruthiventi2016saliency,jetley2016end,cPan,mlnet2016,cornia2017sam}. However, a very large variety of models have been proposed before the advent of deep learning and almost each of them has been inspired by the seminal work of Itti and Koch~\cite{itti1998model}, in which multi-scale low-level features extracted from the input image were linearly combined and then processed by a dynamic neural network with a winner-takes-all strategy. The same idea of properly combining different low-level features was also explored by Harel~\etal\cite{harel2006graph} who defined Markov chains over various image maps, and treated the equilibrium distribution over map locations as an activation. In addition to the exploitation of low-level features, several saliency models have also incorporated high-level concepts such as faces, people, and text~\cite{judd2009learning,borji2012boosting,zhang2013saliency}. In fact, Judd~\etal\cite{judd2009learning} highlighted that, when humans look at images, their gazes are attracted not only by low-level cues typical of the bottom-up attention, but also by top-down image semantics. To this end, they proposed a model in which low and medium level features were effectively combined, and exploited face and people detectors to capture important high-level concepts. Nonetheless, all these techniques have failed to effectively capture the wide variety of causes that contribute to define the visual saliency on images and, with the advent of deep learning, researchers have developed data-driven architectures capable of overcoming many of the limitations of hand-crafted models. 

First attempts of computing saliency maps through a neural network lacked from the absence of sufficiently large training datasets~\cite{vig2014large,kummerer2014deep,liu2015predicting}. Vig~\etal\cite{vig2014large} proposed the first deep architecture for saliency, which was composed by only three convolutional layers. Afterwards, K{\"u}mmerer~\etal~\cite{kummerer2014deep,kummerer2016deepgaze} based their models on two popular convolutional networks (AlexNet~\cite{krizhevsky2012imagenet} and VGG-19~\cite{Simonyan14c}) obtaining adequate results, despite the network parameters were not fine-tuned on a saliency dataset. Liu~\etal\cite{liu2015predicting} tried to overcome the absence of large scale datasets by training their model on image patches centered on fixation and non-fixation locations, thus increasing the amount of training data.      

With the arrival of the SALICON dataset~\cite{jiang2015salicon}, which is still the large publicly available dataset for saliency prediction, several deep architectures have moved beyond previous approaches bringing consistent performance advances. The starting point of all these architectures is a pre-trained Convolutional Neural Network (CNN), such as VGG-16~\cite{Simonyan14c}, GoogleNet~\cite{szegedy2015going} and ResNet~\cite{he2015deep}, to which different saliency-oriented components are added~\cite{mlnet2016,cornia2017sam}, together with different training strategies~\cite{huang2015salicon,jetley2016end,cornia2017sam}. 

In particular, Huang~\etal\cite{huang2015salicon} compared three standard CNNs by applying them at two different image scales. In addition, they were the first to train the network using a saliency evaluation metric as loss function. Jetley~\etal\cite{jetley2016end} introduced a model which formulates a saliency map as generalized Bernoulli distribution. Moreover, they trained their network by using different loss functions which pair the softmax activation function with measures designed to compute distances between probability distributions. Tavakoli~\etal\cite{tavakoli2017exploiting} investigated inter-image similarities to estimate the saliency of a given image using an ensemble of extreme learners, each trained on an image similar to the input image. Kruthiventi~\etal\cite{kruthiventi2016saliency}, instead, presented an unified framework to predict both eye fixations and salient objects. 


Another saliency prediction model was recently presented by Pan~\etal\cite{pan2017salgan} who, following the large dissemination of Generative Adversarial Networks, trained their model by using adversarial examples. In particular, their architecture is composed by two agents: a generator which is responsible for generating the saliency map of a given image, and a discriminator which performs a binary classification task between generated and real saliency maps. Liu~\etal\cite{liu2016deep}, instead, proposed a model to learn long-term spatial interactions and scene contextual modulation to infer image saliency showing promising results, also thanks to the use of the powerful ResNet-50 architecture~\cite{he2015deep}.

In contrast to all these works, we presented two different deep saliency architectures. The first one, called \textit{ML-Net}~\cite{mlnet2016}, effectively combines features coming from different levels of a CNN and applies a matrix of learned weights to the predicted saliency map thus taking into account the center bias present in human eye fixations. The second one, called \textit{SAM}~\cite{cornia2017sam}, incorporates neural attentive mechanisms which focus on the most salient regions of the input image. The core component of the proposed model is an Attentive Convolutional LSTM that iteratively refines the predicted saliency map. Moreover, to tackle the human center bias, the network is able to learn multiple Gaussian prior maps without predefined information. Since this model achieved state of the art performances, being at the top of different saliency prediction benchmarks, we use it in this work.

\subsection{Image captioning}
Recently, the automatic description of images and video has been addressed by computer vision researchers with recurrent neural networks which, given a vectored description of the visual content, can naturally deal with sequences of words~\cite{vinyals2015show,karpathy2015deep,baraldi17cvpr}. Before deep learning models, the generation of sentences was mainly tackled by identifying visual concepts, objects and attributes which were then combined into sentences using pre-defined templates~\cite{yao2010i2t,yang2011corpus,kulkarni2013babytalk}. Another strategy was that of posing the image captioning as a retrieval problem, where the closest annotated sentence in the training set was transferred to a test image, or where training captions were split into parts and then reassembled to form new sentences~\cite{farhadi2010every,ordonez2011im2text,hodosh2013framing,socher2014grounded}. Obviously, all these approaches limited the variety of possible outputs and could not satisfy the richness of natural language. Recent captioning models, in fact, address the generation of sentences as a machine translation problem in which a visual representation of the image coming from a convolutional network is translated in a language counterpart through a recurrent neural network.

One of the first models based on this idea is that proposed by Karpathy~\etal\cite{karpathy2015deep} in which sentence snippets are aligned to the visual regions that they describe through a multimodal embedding. After that, these correspondences are treated as training data for a multimodal recurrent neural network which learns to generate the corresponding sentences. Vinyals~\etal\cite{vinyals2015show}, instead, developed an end-to-end model trained to maximize the likelihood of the target sentence given the input image. Xu~\etal\cite{icml2015xuc15} introduced an approach to image captioning which incorporates a form of machine attention, by which a generative LSTM can focus on different regions of the image while generating the corresponding caption. They proposed two different versions of their model: the first one, called ``Soft Attention'', is trained in a deterministic manner using standard backpropagation techniques, while the second one, called ``Hard Attention'', is trained by maximizing a variational lower bound through the reinforcement learning paradigm.

Johnson~\etal\cite{johnson2016densecap} addressed the task of dense captioning, which jointly localizes and describes in natural language salient image regions. This task consists of generalizing the object detection problem when the descriptions consist of a single word, and the image captioning task when one predicted region covers the full image. You~\etal\cite{you2016image} proposed a semantic attention model in which, given an image, a convolutional neural network extracts top-down visual features and at the same time detects visual concepts such as regions, objects and attributes. The image features and the extracted visual concepts are combined through a recurrent neural network that finally generates the image caption. 
Differently from previous works which aim at predicting a single caption, Krause~\etal\cite{krause2016hierarchical} introduced the generation of entire paragraphs for describing images. Finally, Shetty~\etal\cite{shetty2017speaking} employed adversarial training to change the training objective of the caption generator from reproducing ground-truth captions to generating a set of captions that is indistinguishable from human generated captions.

In this paper, we are interested in demonstrating the importance of using saliency along with contextual information during the generation of image descriptions. Our solution falls in the class of neural attentive captioning architectures and, in the experimental section, we compare it against a standard attentive model built upon the Soft Attention approach presented in~\cite{icml2015xuc15}.

\subsection{Visual saliency and captioning}
Only a few other previous works have investigated the contribution of human eye fixations to generate image descriptions. The first work that has explored this idea was that proposed in~\cite{sugano2016seeing} which presented an extension of a neural attentive captioning architecture. In particular, the proposed model incorporates human fixation points (obtained with eye-tracking devices) instead of computed saliency maps to generate image captions. This kind of strategy mainly suffers of the need of having both eye fixation and caption annotations. Currently, only the SALICON dataset~\cite{jiang2015salicon}, being a subset of the Microsoft COCO dataset~\cite{lin2014microsoft}, is available with both human descriptions and saliency maps. 

Ramanishka~\etal\cite{ramanishka2016top}, instead, introduced an encoder-decoder captioning model in which spatiotemporal heatmaps are produced for predicted captions and arbitrary query sentences without explicit attention layers. They refer to these heatmaps as saliency maps, even though they are internal representations of the network, not related with human attention. Experiments showed that the gain in performance with respect to a standard captioning attentive model is not consistent, even though the computational overhead is lower.

A different approach, presented in~\cite{tavakoli2017can}, explores if image descriptions, by humans or models, agree with saliency and if saliency can benefit image captioning. To this end, they proposed a captioning model in which image features are boosted with the corresponding saliency map by exploiting a moving sliding window and mean pooling as aggregation strategies. Comparisons with respect to a no-saliency baseline did not show significant improvements (especially on the Microsoft COCO dataset).

In this paper, we instead aim at enhancing image captions by directly incorporating saliency maps in a neural attentive captioning architecture. Differently from previous models that exploit human fixation points, we obtain a more general architecture which can be potentially trained using any image captioning dataset, and can predict captions for any input image. In our model, the machine attentive process is split in two different and unrelated paths, one for salient regions and one for context. We demonstrate through extensive experiments that the incorporation of saliency and context can enhance image captioning on different state of art datasets.

\begin{figure}[t]
\centering
\setlength\tabcolsep{0.1pt}
\begin{center}

    \begin{tabular}{m{2.25cm}m{2.25cm}m{2.25cm}m{2.25cm}m{2.25cm}m{2.25cm}}
        \includegraphics[width=0.15\textwidth]{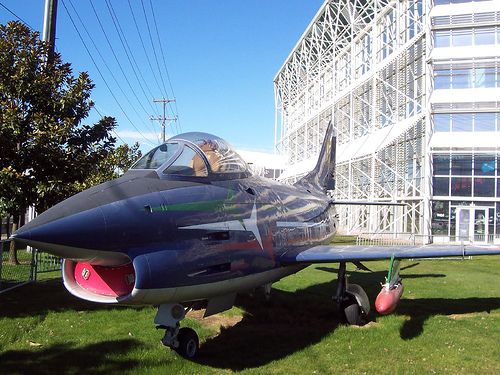}&
        \includegraphics[width=0.15\textwidth]{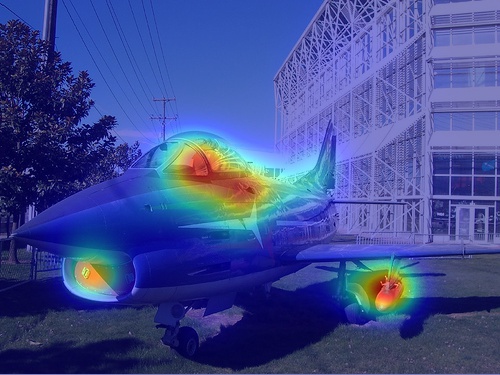}&
        \includegraphics[width=0.15\textwidth]{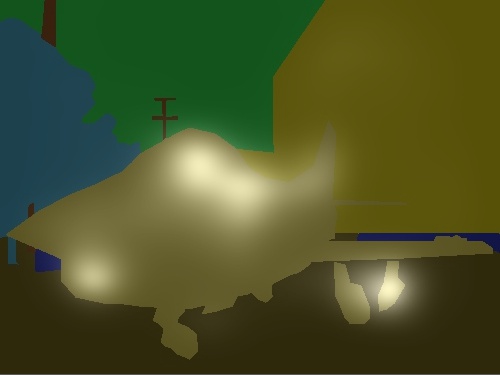}&
        \includegraphics[width=0.15\textwidth]{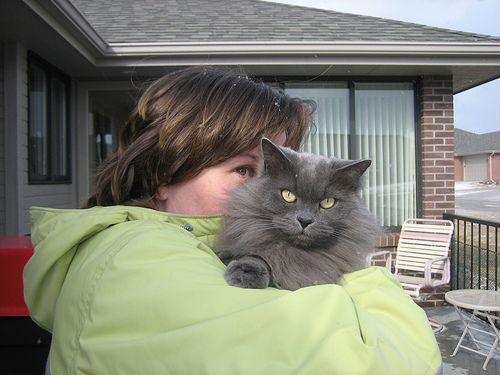}&
        \includegraphics[width=0.15\textwidth]{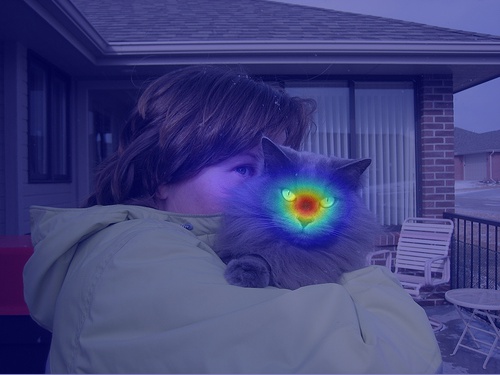}&
        \includegraphics[width=0.15\textwidth]{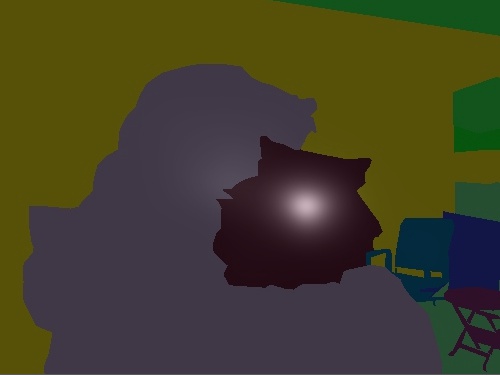}
    \end{tabular}

    \begin{tabular}{m{2.25cm}m{2.25cm}m{2.25cm}m{2.25cm}m{2.25cm}m{2.25cm}}
        \includegraphics[width=0.15\textwidth]{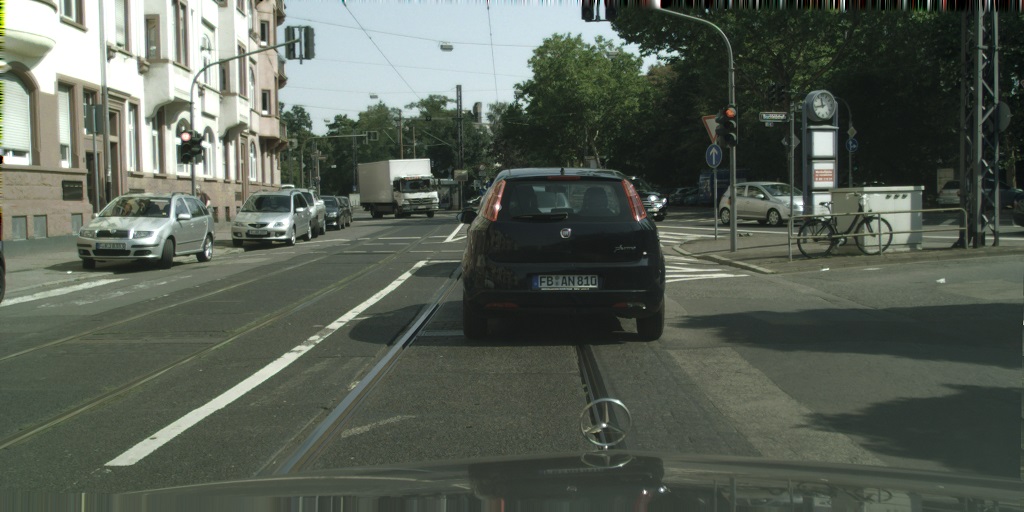}&
        \includegraphics[width=0.15\textwidth]{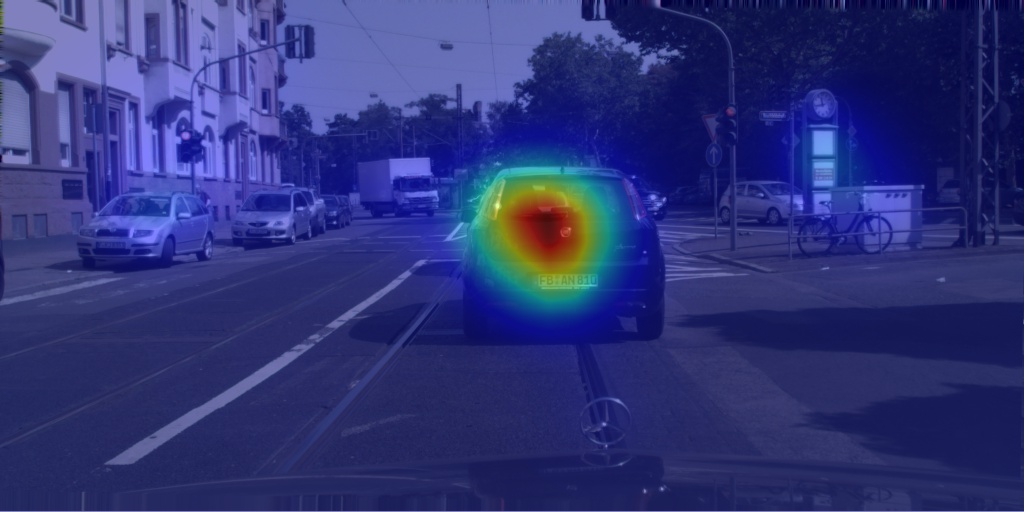}&
        \includegraphics[width=0.15\textwidth]{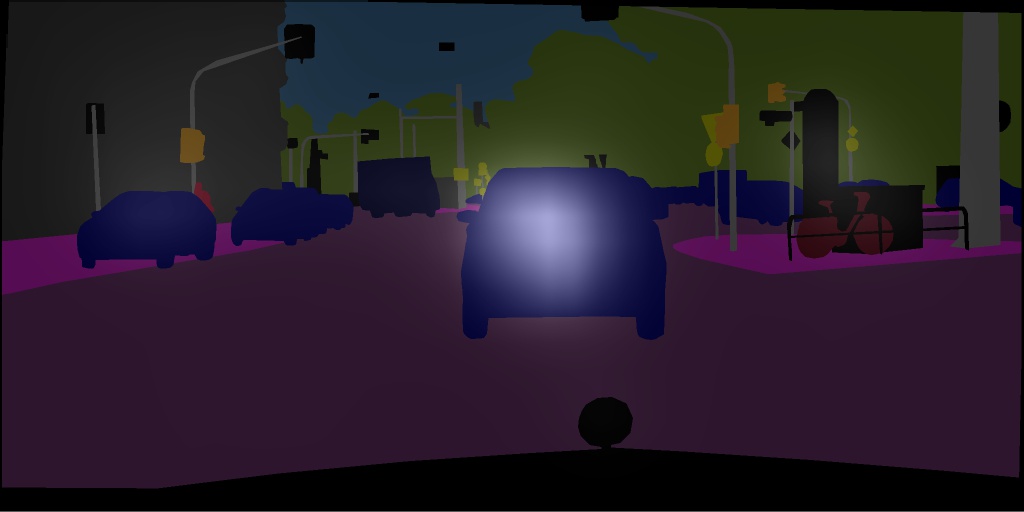}&
        \includegraphics[width=0.15\textwidth]{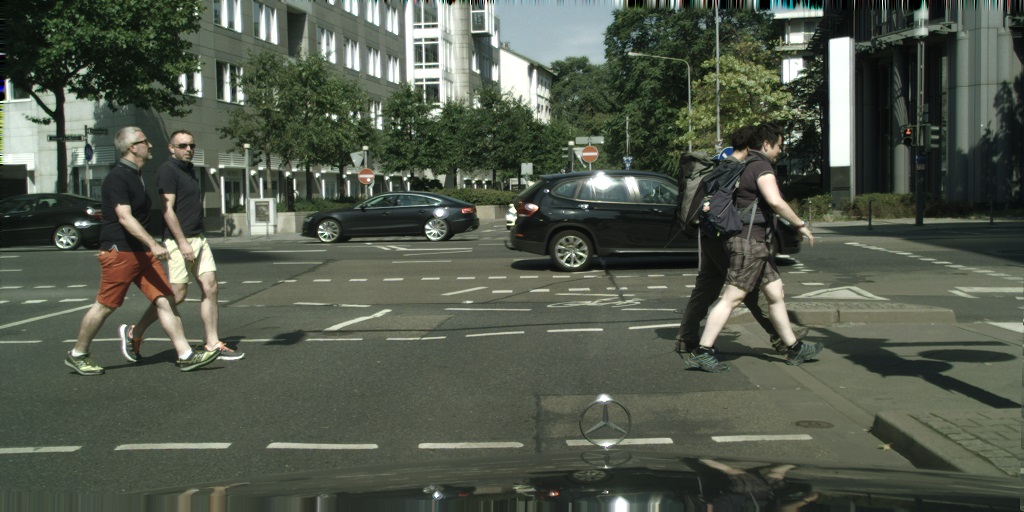}&
        \includegraphics[width=0.15\textwidth]{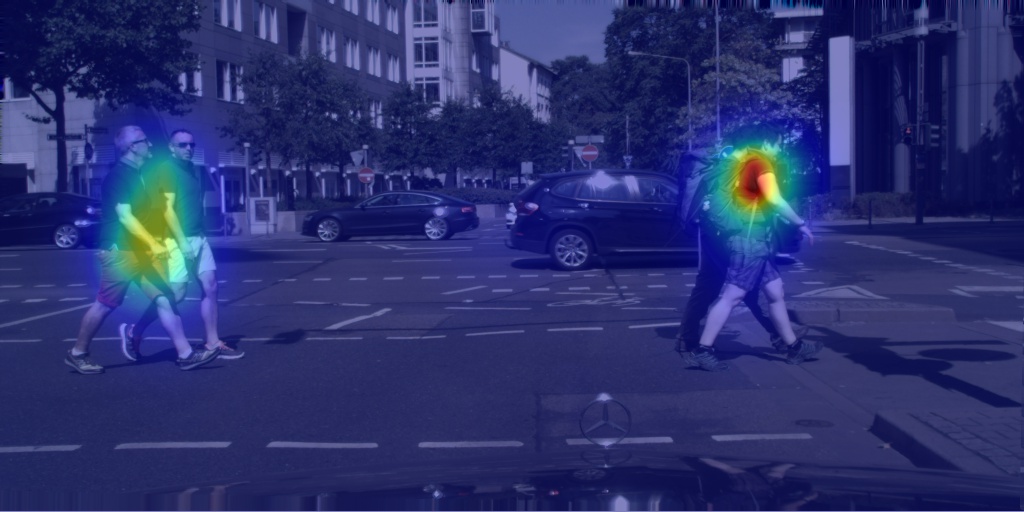}&
        \includegraphics[width=0.15\textwidth]{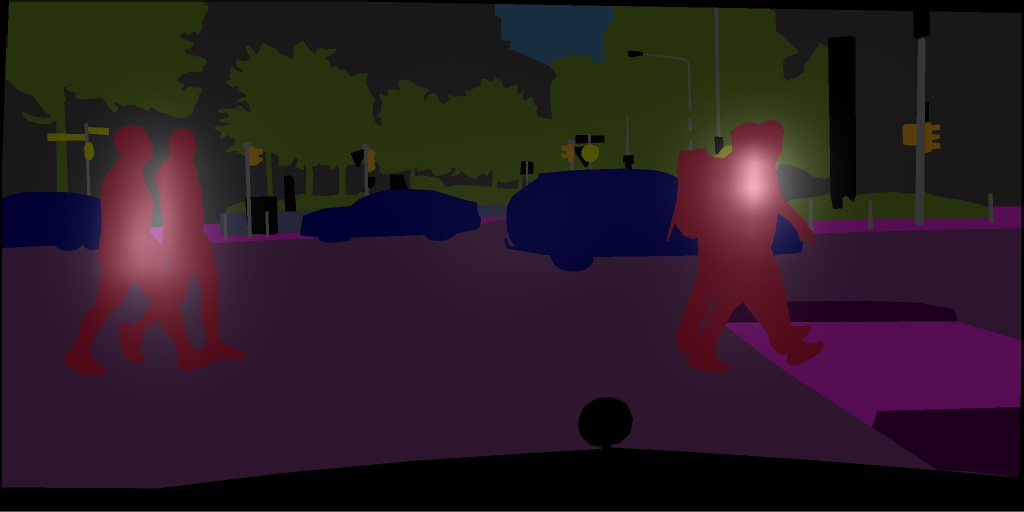}
    \end{tabular}

    \begin{tabular}{m{2.25cm}m{2.25cm}m{2.25cm}m{2.25cm}m{2.25cm}m{2.25cm}}
        \includegraphics[width=0.15\textwidth]{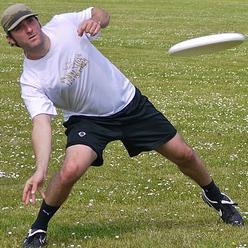}&
        \includegraphics[width=0.15\textwidth]{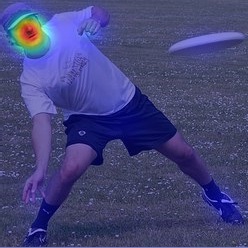}&
        \includegraphics[width=0.15\textwidth]{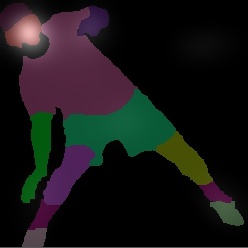}&
        \includegraphics[width=0.15\textwidth]{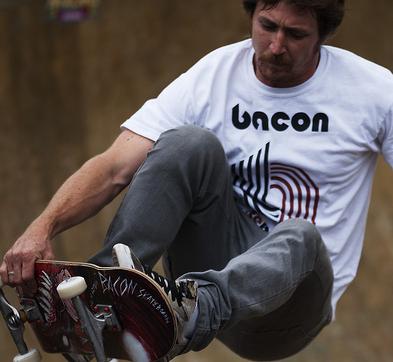}&
        \includegraphics[width=0.15\textwidth]{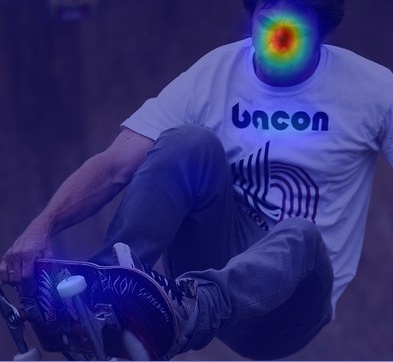}&
        \includegraphics[width=0.15\textwidth]{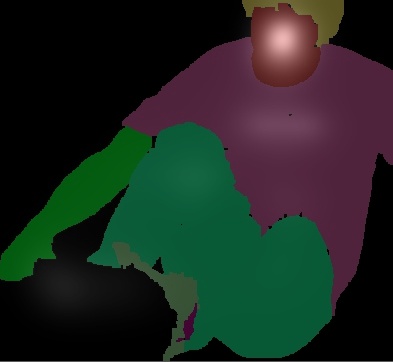}
    \end{tabular}

\end{center}
\caption{Ground-truth semantic segmentation and saliency predictions from our model~\cite{cornia2017sam} on sample images from Pascal-Context~\cite{mottaghi_cvpr14} (first row), Cityscapes~\cite{Cordts2016Cityscapes} (second row) and LIP~\cite{gong2017look} (last row).}
\label{fig:imgs-semantic}
\end{figure}

\section{What is hit by saliency?}
Human gazes are attracted by both low-level cues such as color, contrast and texture, and high-level concepts such as faces and text~\cite{judd2009learning,bylinskii2016should}. Current state of the art saliency prediction methods, thanks to the use of deep networks and large-scale datasets, are able to effectively incorporate all these factors and predict saliency maps which are very close to those obtained from human eye fixations~\cite{cornia2017sam}. In this section we qualitatively investigate which parts of an image are actually hit or ignored by saliency models, by jointly analyzing saliency and semantic segmentation maps. This will motivate the need of using saliency predictions as an additional conditioning for captioning models.

To compute saliency maps, we employ the approach in~\cite{cornia2017sam}, which has shown good results on popular saliency benchmarks, such as the MIT Saliency~\cite{mit-saliency-benchmark} and the SALICON dataset~\cite{jiang2015salicon}, and which also won the LSUN Challenge in 2017. It is worthwhile to mention, anyway, that the qualitative conclusions of this section can be applied to any state of the art saliency model.

Since semantic segmentations algorithms are not always completely accurate, we perform the analysis on three semantic segmentation datasets, in which regions have been segmented by human annotators: Pascal-Context~\cite{mottaghi_cvpr14}, Cityscapes~\cite{Cordts2016Cityscapes} and the Look into Person (LIP)~\cite{gong2017look} dataset. While the first one contains natural images without a specific target, the other two are focused, respectively, on urban streets and human body parts.
In particular, the Pascal-Context provides additional annotations for the Pascal VOC 2010 dataset~\cite{everingham2010pascal} which contains $10,103$ training and validation images and $9,637$ testing images. It goes beyond the original Pascal semantic segmentation task by providing annotations for the whole scene, and images are annotated by using more than $400$ different labels. The Cityscapes dataset, instead, is composed by a set of video sequences recorded in street scenes from $50$ different cities. It provides high quality pixel-level annotations for $5,000$ frames and coarse annotations for $20,000$ frames. The dataset is annotated with $30$ street-specific classes, such as \textit{car}, \textit{road}, \textit{traffic sign}, etc. Finally, the LIP dataset is focused on the semantic segmentation of people and provides more than $50,000$ images annotated with $19$ semantic human part labels. Images contain person instances cropped from the Microsoft COCO dataset~\cite{lin2014microsoft} and split in training, validation and testing sets with $30,462$, $10,000$ and $10,000$ images respectively. For our analyses we only consider train and validation images for the Pascal-Context and LIP datasets, and the $5,000$ pixel-level annotated frames for the Cityscapes dataset. Figure~\ref{fig:imgs-semantic} shows, for some sample images, the predicted saliency map and the corresponding semantic segmentation on the three datasets.

We firstly investigate which are the most and the least salient classes for each dataset. Since there are semantic classes with a low number of occurrences with respect to the total number of images, we only consider relevant semantic classes (\textit{i.e.} classes with at least $N$ occurrences). Due to the different dataset sizes, we set $N$ to $500$ for the Pascal-Context and LIP datasets, and to $200$ for the Cityscapes dataset. To collect the number of times that the predicted saliency hits a semantic class, we binarize each map by thresholding the values of its pixels. A low threshold value leads to a binarized map with dilated salient regions, while an high threshold creates small salient regions around the fixation points. For this reason, we use two different threshold values to analyze the most and the least salient classes. We choose a threshold near $0$ to find the least salient classes for each dataset, and a value near $255$ to find instead the most salient ones.

\begin{figure}[tb]
\centering
\subfloat[Pascal-Context~\cite{mottaghi_cvpr14}] {
	\includegraphics[width=0.25\textwidth,valign=t]{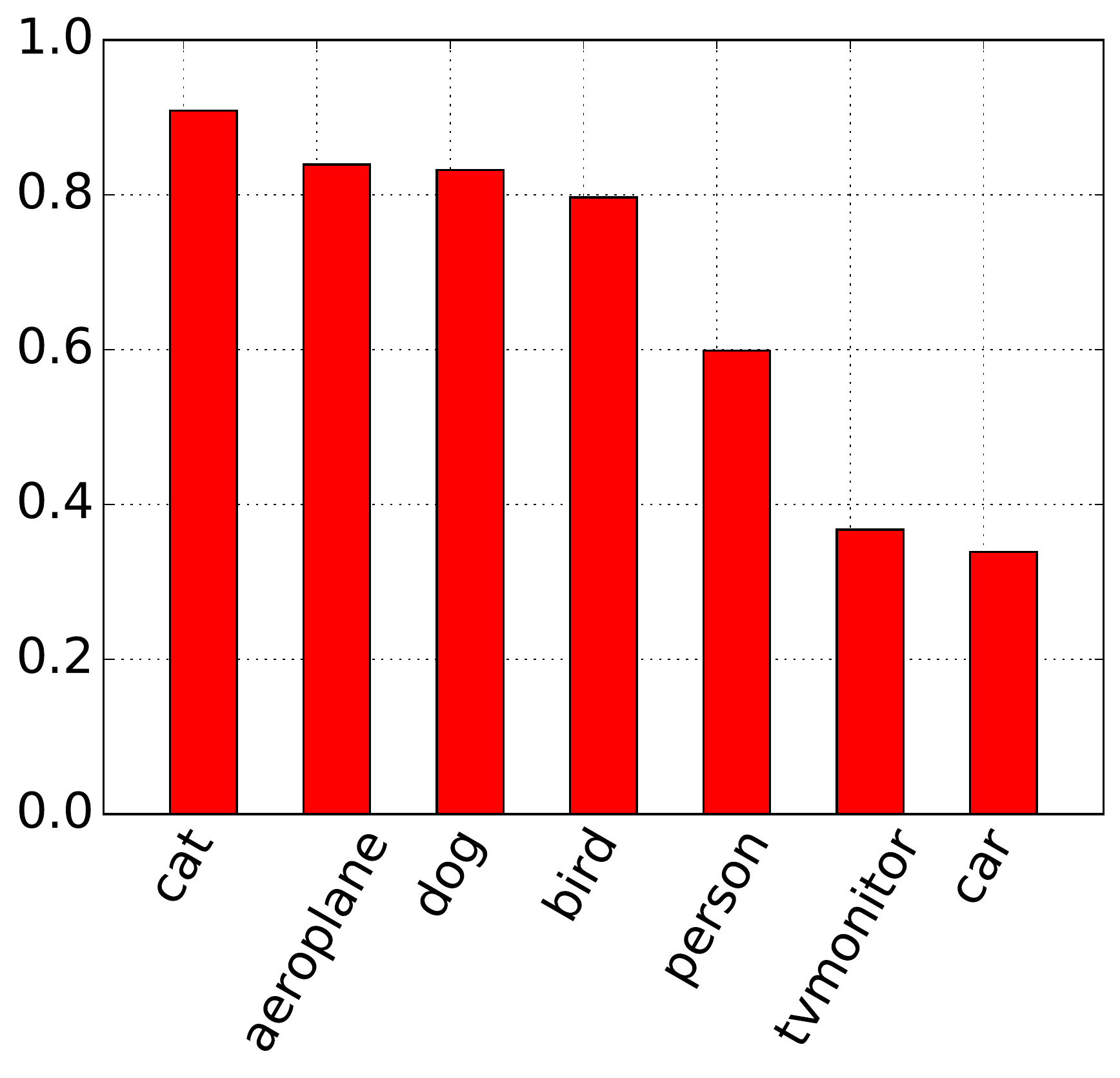}
	\vphantom{\includegraphics[width=0.25\textwidth,valign=t]{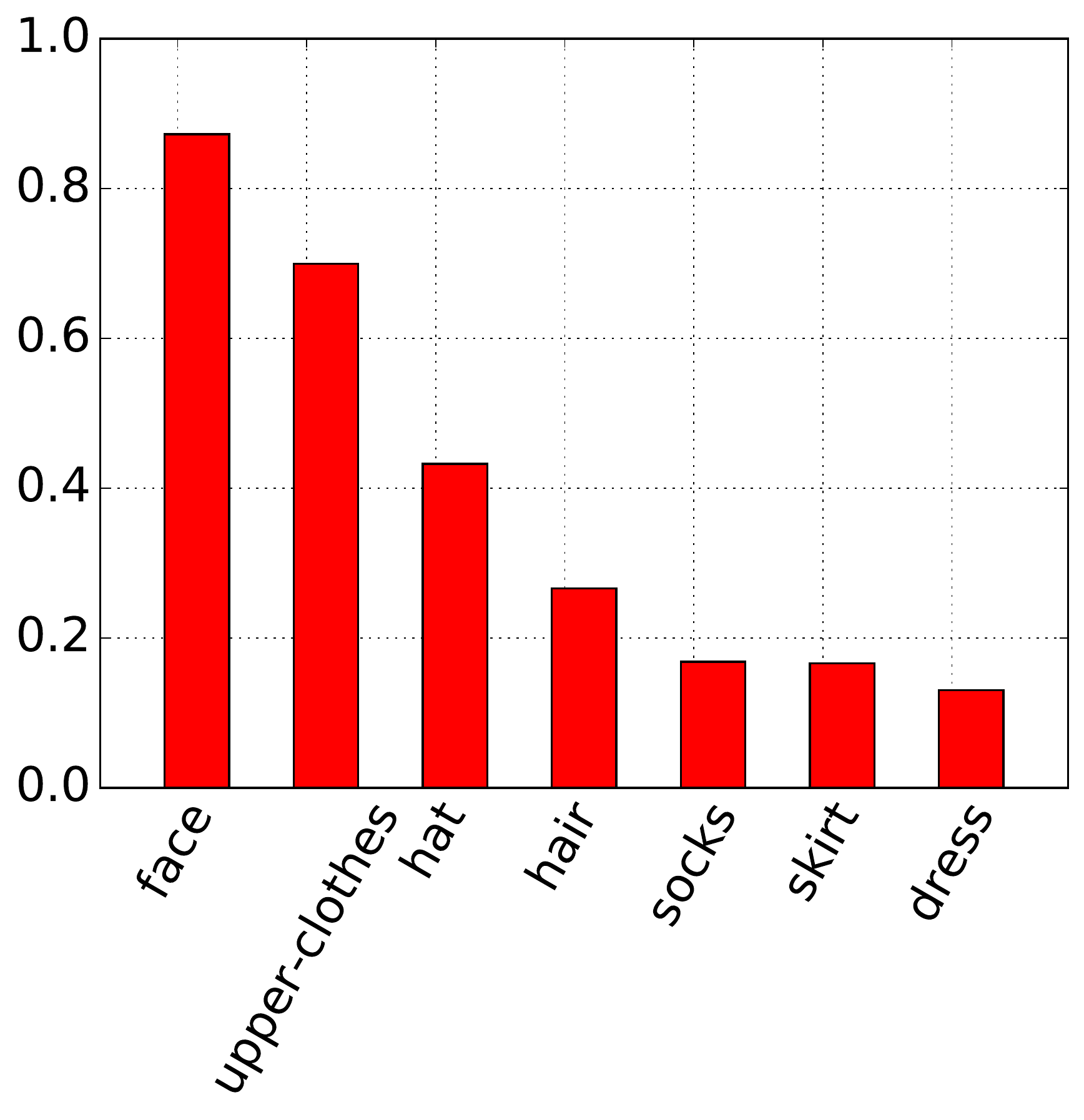}}
}
\subfloat[Cityscapes~\cite{Cordts2016Cityscapes}] {
	\includegraphics[width=0.25\textwidth,valign=t]{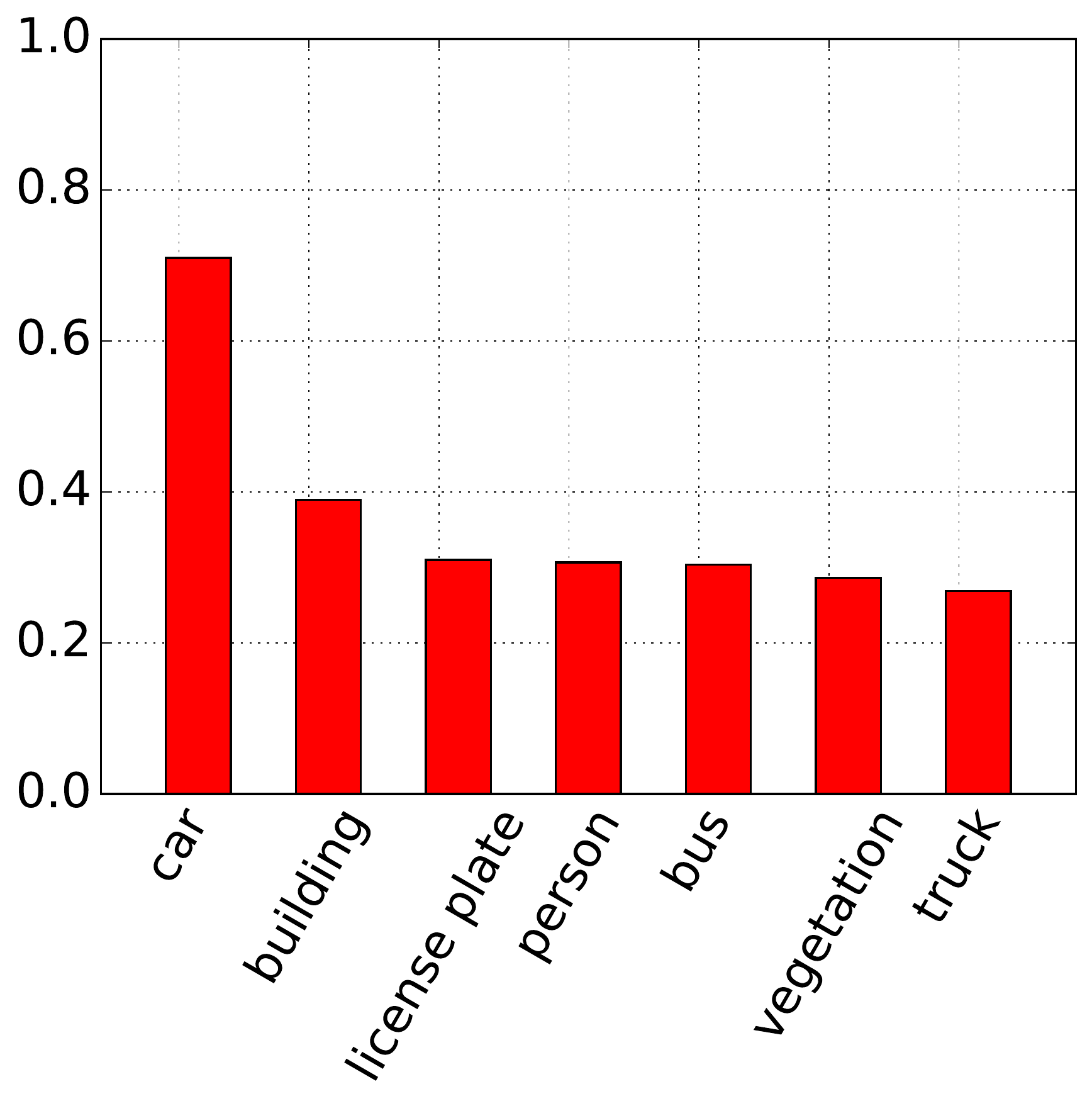}
	\vphantom{\includegraphics[width=0.25\textwidth,valign=t]{images/most_salient_lip.pdf}}
}
\subfloat[LIP~\cite{gong2017look}] {
	\includegraphics[width=0.25\textwidth,valign=t]{images/most_salient_lip.pdf}
}
\caption{Most salient classes on Pascal-Context, Cityscapes and LIP datasets.}
\label{fig:most}
\end{figure}

\begin{figure}[tb]
\centering
\subfloat[Pascal-Context~\cite{mottaghi_cvpr14}] {
	\includegraphics[width=0.25\textwidth,valign=t]{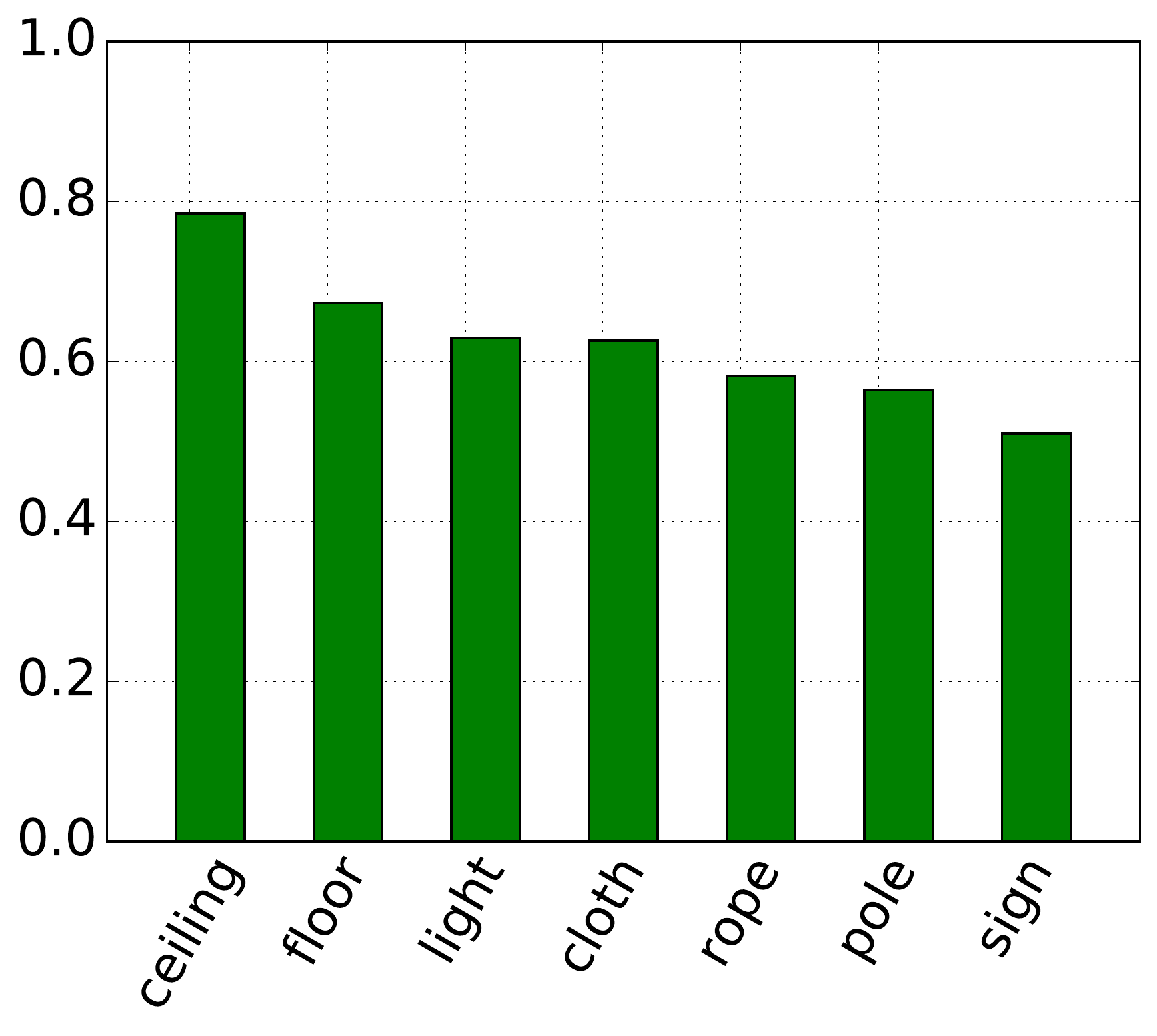}
	\vphantom{\includegraphics[width=0.25\textwidth,valign=t]{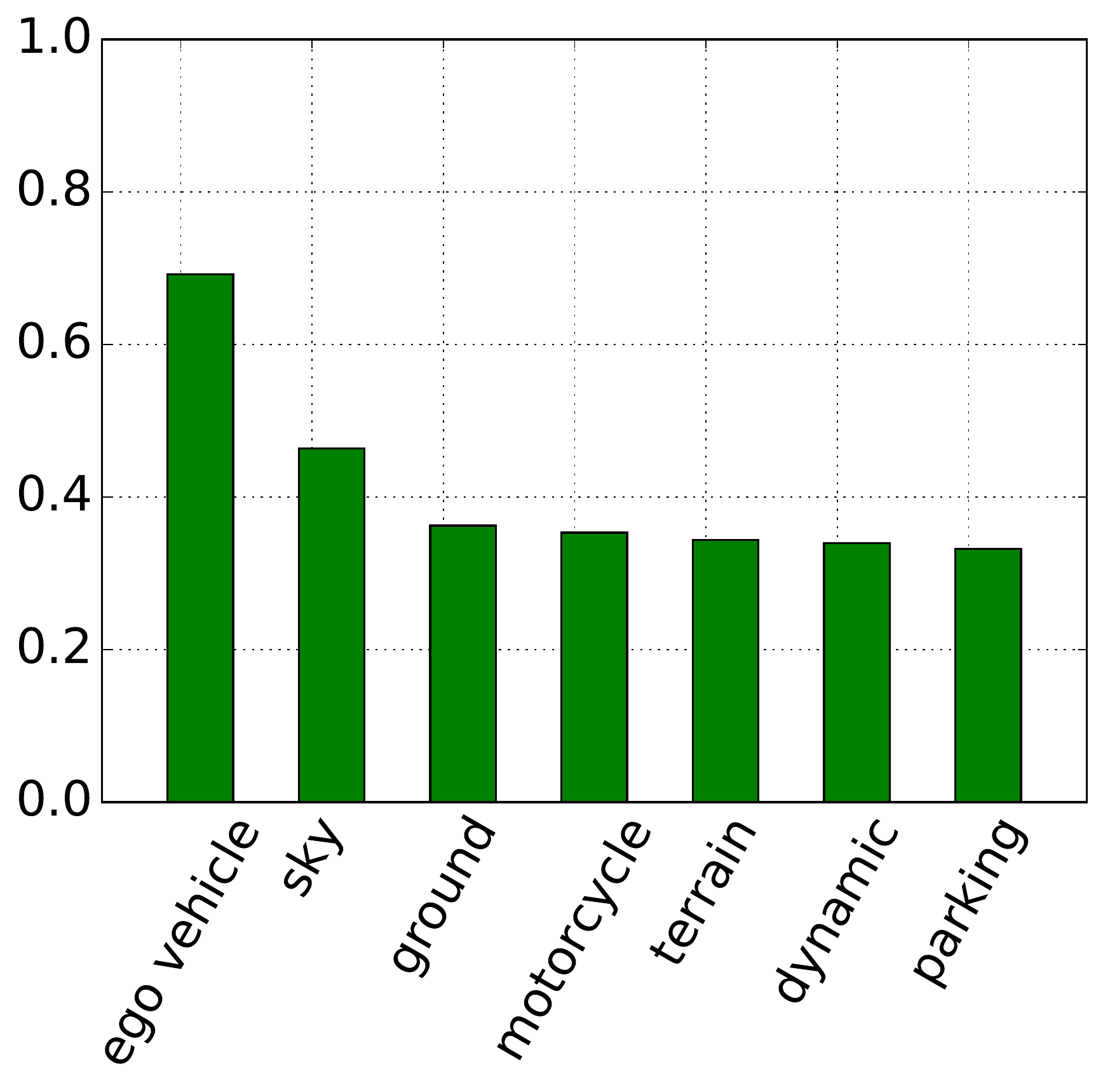}}
}
\subfloat[Cityscapes~\cite{Cordts2016Cityscapes}] {
	\includegraphics[width=0.25\textwidth,valign=t]{images/less_salient_cityscapes.pdf}
}
\subfloat[LIP~\cite{gong2017look}] {
	\includegraphics[width=0.25\textwidth,valign=t]{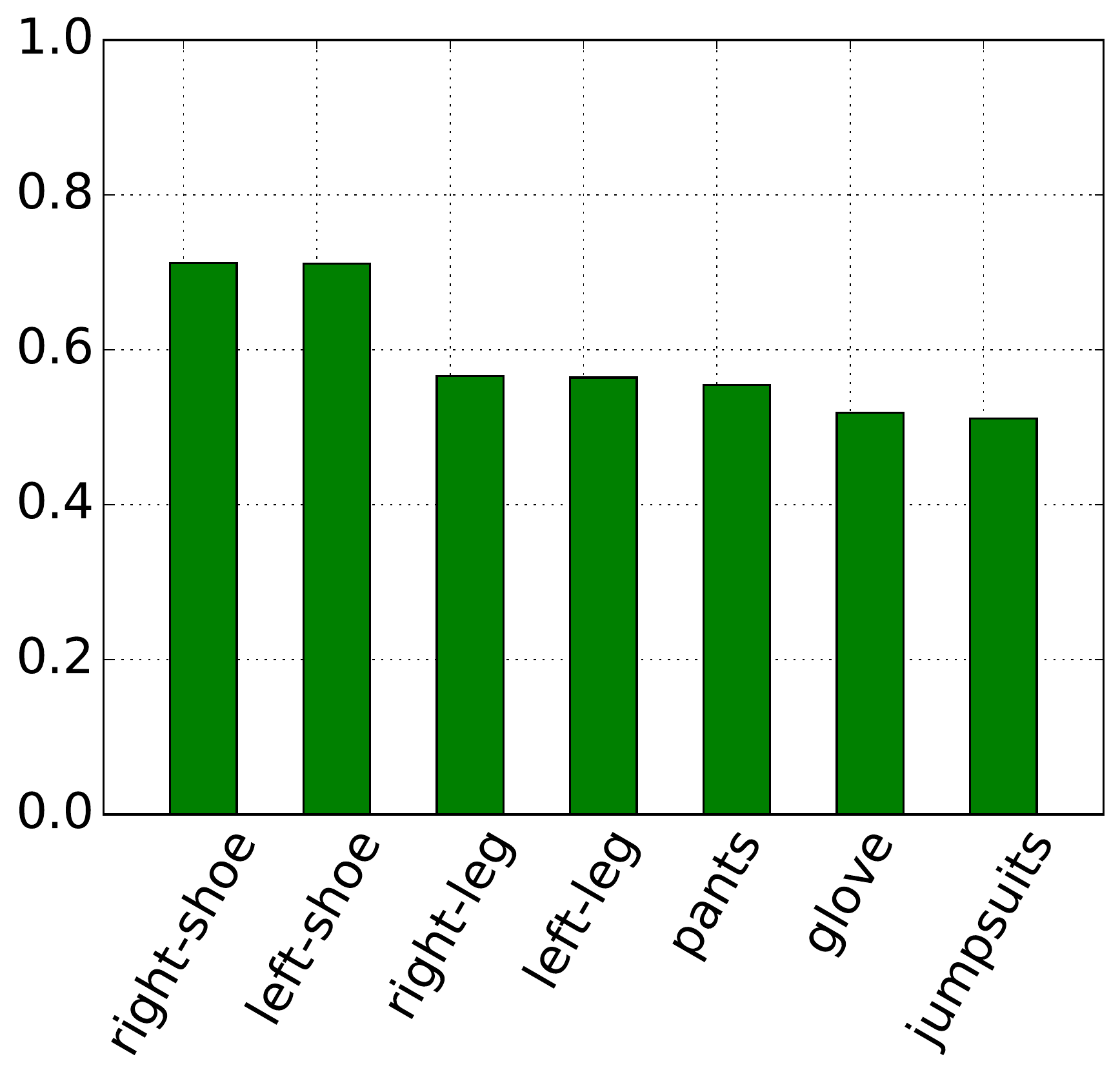}
		\vphantom{\includegraphics[width=0.25\textwidth,valign=t]{images/less_salient_cityscapes.pdf}}
}
\caption{Least salient classes on Pascal-Context, Cityscapes and LIP datasets.}
\label{fig:less}
\end{figure}

Figures~\ref{fig:most} and~\ref{fig:less} show the most and the least salient classes in terms of the percentage of times that saliency hits a region belonging to a class. As it can be seen, there are different distributions depending on the considered dataset. For example, for the Pascal-Context, the most salient classes are animals (such as cats, dogs and birds), people and vehicles (such as airplanes and cars), while the least salient ones result to be ceiling, floor and light. As for the Cityscapes dataset, cars are absolutely the most salient class with a $70\%$ of times in which is hit by saliency. All other classes, instead, do not reach the $40\%$. On the LIP dataset, the most salient classes are all human body parts in the upper body, while the least salient ones are all in the lower body. As expected, people faces are those most hit by saliency with an absolute number of occurrences near to $90\%$. It can be observed as a general pattern that \emph{the most important or visible objects in the scene are hit by saliency, while objects in the background, and the context itself of the image are usually ignored}. This leads to the hypothesis that both salient and non salient regions are important to generate the description of an image, given that we generally want the context to be included in the caption, and that the distinction between salient regions and context, given by a saliency prediction model, can improve captioning results.

\begin{figure}[t!]
\centering
\subfloat[Pascal-Context~\cite{mottaghi_cvpr14}] {
	\includegraphics[width=0.25\textwidth]{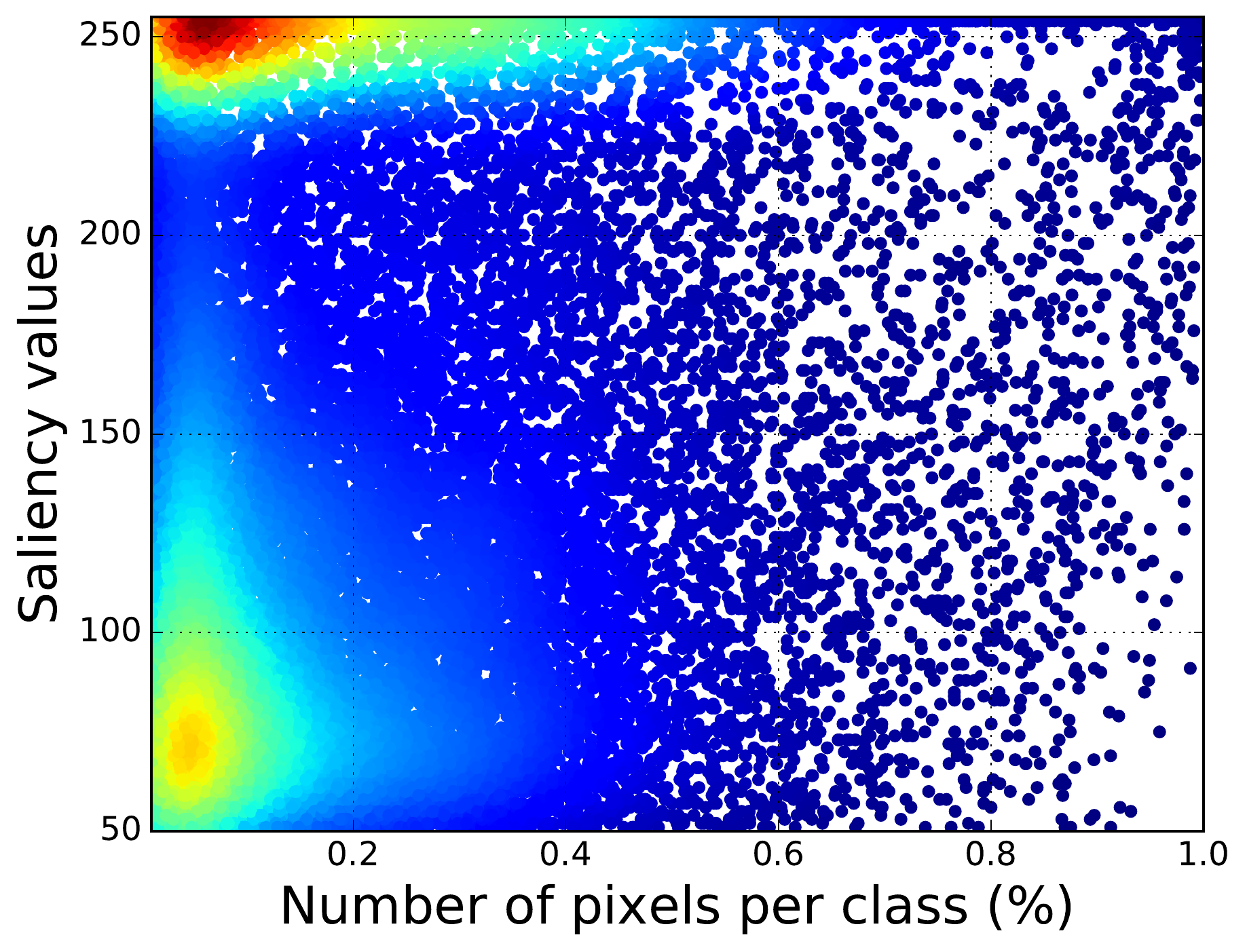}
}
\subfloat[Cityscapes~\cite{Cordts2016Cityscapes}] {
	\includegraphics[width=0.25\textwidth]{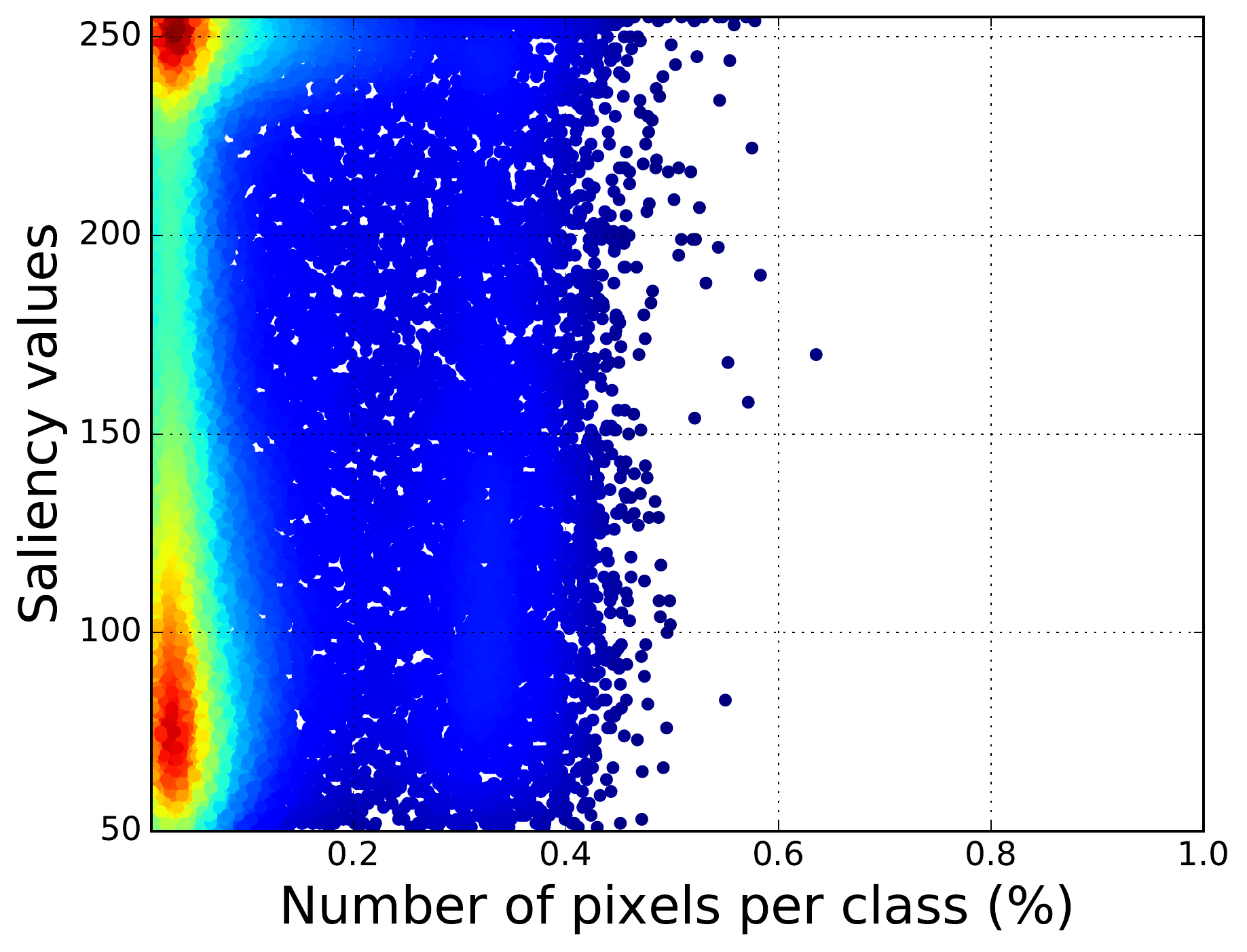}
}
\subfloat[LIP~\cite{gong2017look}] {
	\includegraphics[width=0.25\textwidth]{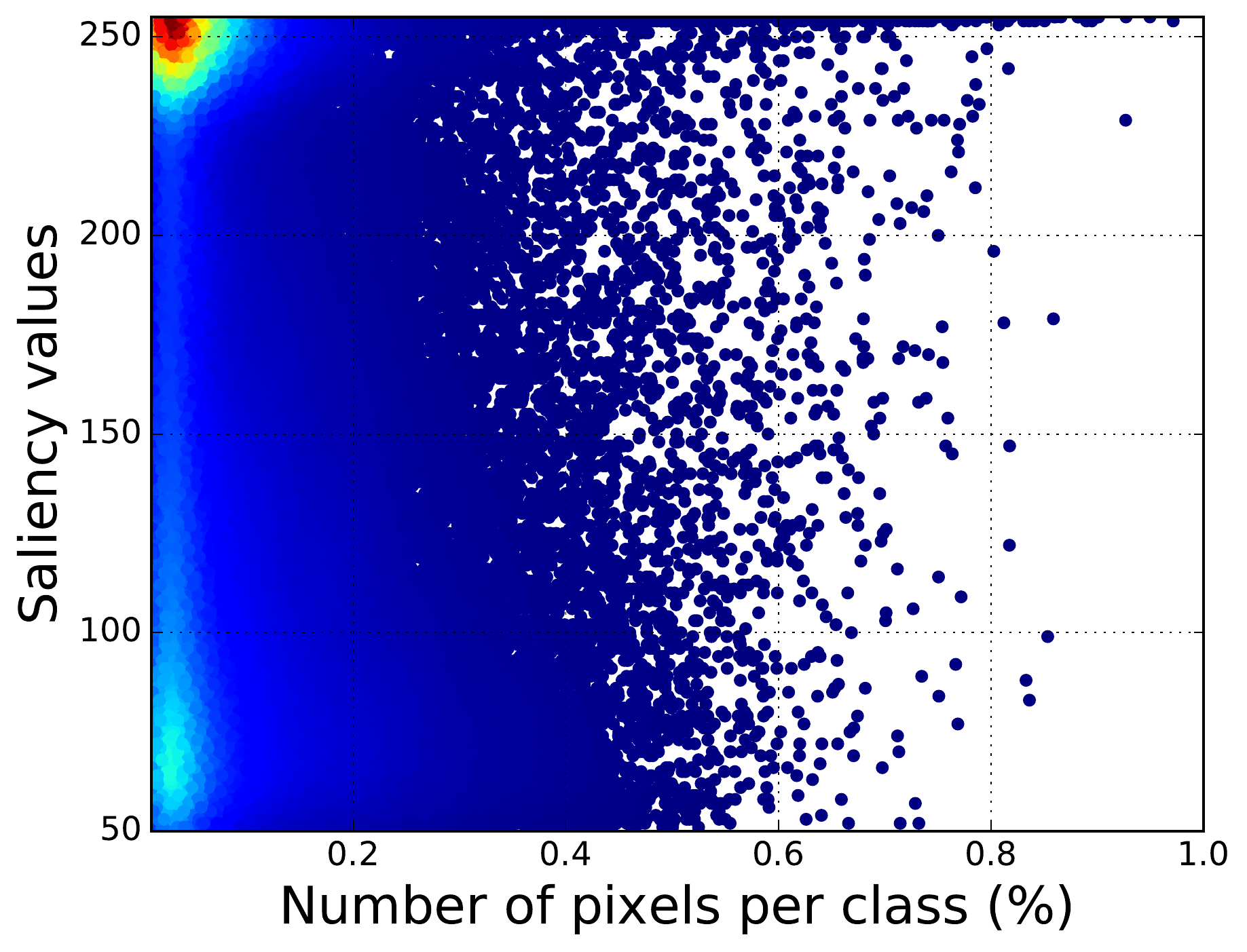}
}
\caption{Distribution of object sizes and saliency values (best seen in color).}
\label{fig:areas}
\end{figure}

We also investigate the existence of a relation between the size of an object and its saliency values. In Figure~\ref{fig:areas}, we plot the joint distribution of object sizes and saliency values on the three datasets, where the size of an object is simply computed as the number of its pixels normalized by the size of the image. As it can be seen, most of the low saliency instances are small; however, high saliency values concentrate on small objects as well as on large ones. In summary, \emph{there is not always a proportionality between the size of an object and its saliency}, so the importance of an object can not be assessed by simply looking at its size. In the image captioning scenario that we want to tackle, larger objects correspond to larger activations in the last layers of a convolutional architecture, while smaller objects correspond to smaller activations. Since salient and non salient regions can have comparable activations, the supervision given by a saliency prediction model on whether a pixel belongs or not to a salient region can be beneficial during the generation of the caption.

\section{Saliency and Context Aware Attention}
\begin{figure}[t!]
\centering
	\includegraphics[width=0.95\textwidth]{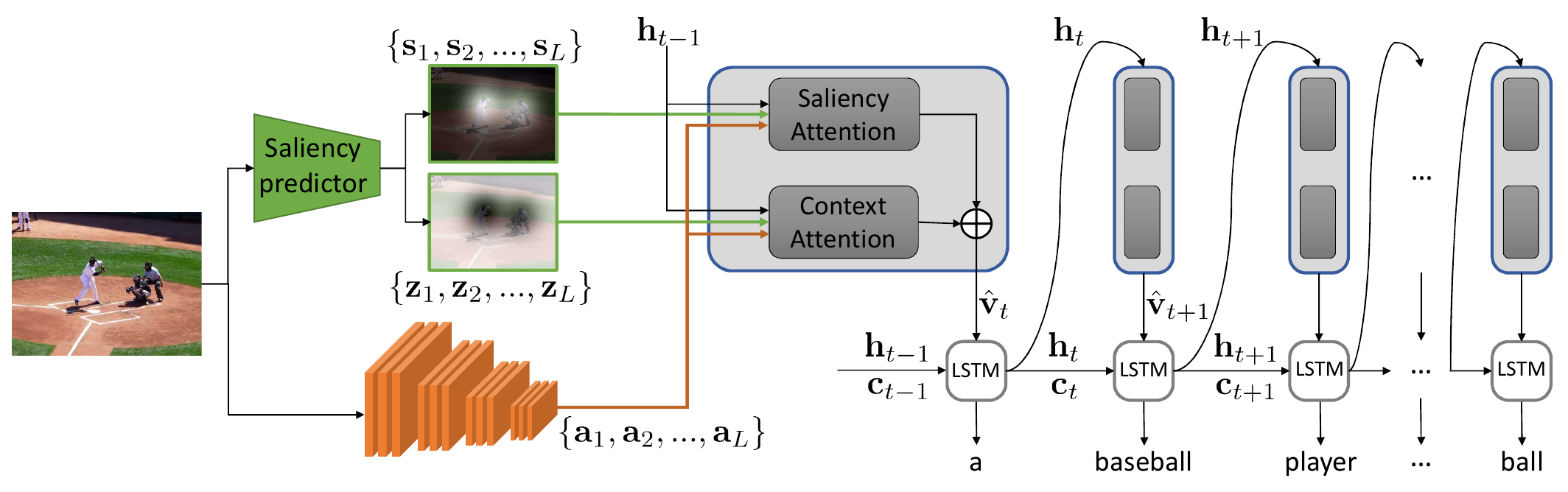}
\caption{Overview of the proposed model. Two different attention paths are built for salient regions and contextual regions, to help the model build captions which describe both components (best seen in color).}
\label{fig:model}
\end{figure}

Following the qualitative findings of the previous section, we develop a model in which saliency is exploited to enhance image captioning. Here, a generative recurrent neural network is conditioned, step by step, on salient spatial regions, predicted by a saliency model, and on contextual features which account for the role on non-salient regions in the generation of the caption. In the following, we describe the overall model. An overview is presented in Figure~\ref{fig:model}.

Each input image $I$ is firstly encoded through a Fully Convolutional Network, which provides a stack of high-level features on a spatial grid $\{\mathbf{a}_1, \mathbf{a}_2, ..., \mathbf{a}_L\}$, each corresponding to a spatial location of the image. At the same time, we extract a saliency map for the input image using the model in~\cite{cornia2017sam}, and downscale it to fit the spatial size of the convolutional features, so to obtain a spatial grid $\{s_1, s_2, ..., s_L\}$ of salient regions, where $s_i \in \left[ 0, 1 \right]$. Correspondingly, we also define a spatial grid of contextual regions, $\{z_1, z_2, ..., z_L\}$ where $z_i = 1-s_i$. Under the model, visual features at different locations will be selected or inhibited according to their saliency value.

The generation of the caption is carried out word-by-word by feeding and sampling words from an LSTM layer, which, at every timestep, is conditioned on features extracted from the input image and on the saliency map. Formally, the behaviour of the generative LSTM is driven by the following equations:
\begin{align}
\mathbf{i}_t &= \sigma(W_{vi} \hat{\mathbf{v}}_t + W_{wi} \mathbf{w}_t + W_{hi} \mathbf{h}_{t-1} + \mathbf{b}_i) \\
\mathbf{f}_t &= \sigma(W_{vf} \hat{\mathbf{v}}_t + W_{wf} \mathbf{w}_t + W_{hf} \mathbf{h}_{t-1} + \mathbf{b}_f) \\
\mathbf{o}_t &= \sigma(W_{vo} \hat{\mathbf{v}}_t + W_{wo} \mathbf{w}_t + W_{ho} \mathbf{h}_{t-1} + \mathbf{b}_o) \\
\mathbf{g}_t &= \phi(W_{vg} \hat{\mathbf{v}}_t + W_{wg} \mathbf{w}_t + W_{hg} \mathbf{h}_{t-1} + \mathbf{b}_g) \\
\mathbf{c}_t &= \mathbf{f}_t \odot \mathbf{c}_{t-1} + \mathbf{i}_t \odot \mathbf{g}_t \\
\mathbf{h}_t &= \mathbf{o}_t \odot \phi(\mathbf{c}_t)
\label{eq:lstm}
\end{align}
where, at each timestep, $\hat{\mathbf{v}}_t$ denotes the visual features extracted from $I$, by considering the map of salient regions $\{ s_i \}_i$, and those of contextual regions $\{ z_i \}_i$. $\mathbf{w}_t$ is the input word, and $\mathbf{h}$ and $\mathbf{c}$ are respectively the internal state and the memory cell of the LSTM. $\odot$ denotes the element-wise Hadamard product, $\sigma$ is the sigmoid function, $\phi$ is the hyperbolic tangent \texttt{tanh}, $W_{*}$ are learned weight matrices and $\mathbf{b}_*$ are learned biases vectors.

To provide the generative network with visual features, we draw inspiration from the machine attention literature~\cite{icml2015xuc15} and compute the fixed-length feature vector $\hat{\mathbf{v}}_t$ as a linear combination of spatial features $\{\mathbf{a}_1, \mathbf{a}_2, ..., \mathbf{a}_L\}$ with time-varying weights $\alpha_{ti}$, normalized over the spatial extent via a \textit{softmax} operator:
\begin{align}
    \hat{\mathbf{v}}_{t} &= \sum_{i=1}^{L} \alpha_{ti} \mathbf{a}_i,\label{eq:att_sum} \\
    \alpha_{ti} &= \frac{\exp{(e_{ti})}}{\sum_{k=1}^{L} \exp{(e_{tk})}}.
\end{align}
At each timestep the attention mechanism selects a region of the image, based on the previous LSTM state, and feeds it to the LSTM, so that the generation of a word is conditioned on that specific region, instead of being driven by the entire image.

Ideally, we want weights $\alpha_{ti}$ to be aware of the saliency and contextual value of location $\mathbf{a}_i$, and to be conditioned on the current status of the LSTM, which can be well encoded by its internal state $\mathbf{h}_t$. In this way, the generative network can focus on different locations of the input image according to their belonging to a salient or contextual region, and to the current generation state. Of course, simply multiplying attention weights with saliency values would result in a loss of context, which is fundamental for caption generation. We instead split attention weights $e_{ti}$ into two contributions, one for saliency and one for context regions, and employ two different fully connected networks to learn the two contributions (Figure~\ref{fig:model}). Conceptually, this is equivalent to building two separate attention paths, one for salient regions and for contextual regions, which are merged to produce the final attention. Overall, the model obeys to the following equation:
\begin{equation}
e_{ti} = s_i \cdot e_{ti}^{sal} + z_i \cdot e_{ti}^{ctx}  \label{eq:att2}
\end{equation}
where $e_{ti}^{sal}$ and $e_{ti}^{ctx}$ are, respectively, the attention weights for salient and context regions. Attention weights for saliency and context are computed as follows:
\begin{align}
    e_{ti}^{sal} &= v_{e, sal}^T \cdot \phi (W_{ae, sal} \cdot \mathbf{a}_i + W_{he, sal} \cdot \mathbf{h}_{t-1}) \\
    e_{ti}^{ctx} &= v_{e, ctx}^T \cdot \phi (W_{ae, ctx} \cdot \mathbf{a}_i + W_{he, ctx} \cdot \mathbf{h}_{t-1})
\end{align}
Notice that our model learns different weights for saliency and contextual regions, and combines them into a final attentive map in which the contributions of salient and non-salient regions are merged together. Similarly to the classical Soft Attention approach~\cite{icml2015xuc15}, the proposed generative LSTM can focus on every region of the image, but the attentive process is aware of the saliency of each location, so that the focus on salient and contextual regions is driven by the output of the saliency predictor.

\subsection{Sentence generation}
Words are encoded with one-hot vectors having size equal to that of the vocabulary, and are then projected into an embedding space via a learned linear transformation. Because sentences have different lengths, they are also marked with special begin-of-string and end-of-string tokens, to keep the model aware of the beginning and end of a particular sentence.

Given an image and sentence $(\mathbf{y}_0, \mathbf{y}_1, ..., \mathbf{y}_T)$, encoded with one-hot vectors, the generative LSTM is conditioned step by step on the first $t$ words of the caption, and is trained to produce the next word of the caption. The objective function which we optimize is the log-likelihood of correct words over the sequence
\begin{equation}
\max_{\mathbf{w}} \sum_{t=1}^T \log \text{Pr} (\mathbf{y}_t | \hat{\mathbf{v}}_t, \mathbf{y}_{t-1}, \mathbf{y}_{t-2}, ..., \mathbf{y}_0)
\end{equation}
where $\mathbf{w}$ are all the parameters of the model. The probability of a word is modeled via a softmax layer applied on the output of the LSTM. To reduce the dimensionality, a linear embedding transformation is used to project one-hot word vectors into the input space of the LSTM and, viceversa, to project the output of the LSTM to the dictionary space.
\begin{equation}
\text{Pr} (\mathbf{y}_t | \hat{\mathbf{v}}_t, \mathbf{y}_{t-1}, \mathbf{y}_{t-2}, ..., \mathbf{y}_0) \propto \exp (\mathbf{y}_t^T W_p \mathbf{h}_t)
\end{equation}
where $W_p$ is a matrix for transforming the LSTM output space to the word space and $\mathbf{h}_t$ is the output of the LSTM.

At test time, the LSTM is given a begin-of-string tag as input for the first timestep, then the most probable word according to the predicted distribution is sampled and given as input for the next timestep, until an end-of-string tag is predicted.

\section{Experimental evaluation}
In this section we perform qualitative and quantitative experiments to validate the effectiveness of the proposed model with respect to different baselines and other saliency-boosted captioning methods. First, we describe datasets and metrics used to evaluate our solution and provide implementation details. 

\subsection{Datasets and metrics}
\label{sec:dataset_metrics}
To validate the effectiveness of the proposed Saliency and Context aware Attention, we perform experiments on five popular image captioning datasets: SALICON~\cite{jiang2015salicon}, Microsoft COCO~\cite{lin2014microsoft}, Flickr8k~\cite{hodosh2013framing}, Flickr30k~\cite{young2014image}, and PASCAL-50S~\cite{vedantam2015cider}.

Microsoft COCO is composed by more than $120,000$ images divided in training and validation sets, where each of them is provided with at least five sentences generated by using Amazon Mechanical Turk. SALICON is a subset of this one, created for the visual saliency prediction task. Since its images are from the Microsoft COCO dataset, at least five captions for each image are available. Overall, it contains $10,000$ training images, $5,000$ validation images and $5,000$ testing images where eye fixations for each image are simulated with mouse movements. In our experiments, we only use train and validation sets for both datasets. The Flickr8k and the Flickr30k datasets are composed by $8,000$ and $30,000$ images respectively. Both of them come with five annotated sentences for each image. In our experiments, we randomly choose $1,000$ validation images and $1,000$ test images for each of these two datasets. The PASCAL-50S dataset provides additional annotations for the UIUC PASCAL sentences~\cite{rashtchian2010collecting}. It is composed of $1,000$ images from the PASCAL-VOC dataset, each of them annotated with $50$ human-written sentences, instead of $5$ as in the original dataset. Due to the limited number of samples and for a fair comparison with other captioning methods, we first pre-train the model on the Microsoft COCO dataset, then we test it on the images of this dataset without a specific fine-tuning.

For evaluation, we employ four automatic metrics which are usually employed in image captioning: BLEU~\cite{papineni2002bleu}, $\text{ROUGE}_L$~\cite{lin2004rouge}, METEOR~\cite{banerjee2005meteor} and CIDEr~\cite{vedantam2015cider}. 
BLEU is a modified form of precision between n-grams to compare a candidate translation against multiple reference translations. We evaluate our predictions with BLEU using mono-grams, bi-grams, three-grams and four-grams. 
$\text{ROUGE}_L$ computes an F-measure considering the longest co-occurring in sequence n-grams.
METEOR, instead, is based on the harmonic mean of unigram precision and recall, with recall weighted higher than precision. It also has several features that are not found in other metrics, such as stemming and synonymy matching, along with the standard exact word matching. CIDEr, finally, computes the average cosine similarity between n-grams found in the generated caption and those found in reference sentences, weighting them using TF-IDF. To ensure a fair evaluation, we use the Microsoft COCO evaluation toolkit\footnote{\url{https://github.com/tylin/coco-caption}} to compute all scores.

\subsection{Implementation details}
Each image is encoded through a convolutional network, which computes a stack of high-level features. We employ the popular ResNet-50~\cite{he2015deep}, trained over the ImageNet dataset~\cite{russakovsky2015imagenet}, to compute the feature maps over the input image. In particular, the ResNet-50 is composed by 49 convolutional layers, divided in 5 convolutional blocks, and 1 fully connected layer. Since we want to maintain the spatial dimensions, we extract the feature maps from the last convolutional layer and ignore the fully connected layer. The output of the ResNet model is a tensor with $2048$ channels. To limit the number of feature maps and the number of learned parameters, we fed this tensor into another convolutional layer with $512$ filters and a kernel size of $1$, followed by a ReLU activation function. Differently from the weights of the ResNet-50 which are kept fixed, the weights of this last convolutional layer are initialized according to~\cite{glorot2010understanding} and fine-tuned over the considered datasets. In the LSTM, following the initialization proposed in~\cite{bahdanau2014neural}, the weight matrices applied to the inputs are initialized by sampling each element from the Gaussian distribution of $0$ mean and $0.01^2$ variance, while the weight matrices applied to the internal states are initialized by using the orthogonal initialization. The vectors $v_{e, sal}$ and $v_{e, ctx}$ as well as all bias vectors $\mathbf{b}_*$ are instead initialized to zero. 

To predict the saliency map for each input image, we exploit our Saliency Attentive Model (SAM)~\cite{cornia2017sam} which is able to predict accurate saliency maps according to different saliency benchmarks. We note however, that we do not expect a significant performance variation when using other state of the art saliency methods.

\begin{table}[t]
    \begin{center}
    \renewcommand{\arraystretch}{1.2}
    \caption{Image captioning results. The conditioning of saliency and context (\texttt{Saliency+Context Attention}) enhances the generation of the caption with respect to the traditional machine attention mechanism. \texttt{Soft Attention} here indicates our reimplementation of~\cite{icml2015xuc15}, using the same visual features of our model.}
    \label{tab:captioning}
    \begin{small}
    \begin{tabular}{|c|c|ccccccc|}
    \hline
    \footnotesize{Dataset} 	& \footnotesize{Model} & \footnotesize{B@1} & \footnotesize{B@2} &  \footnotesize{B@3} & \footnotesize{B@4} & \footnotesize{METEOR} &  \footnotesize{ROUGE$_L$} &  \footnotesize{CIDEr} \\ \hline \hline 
    \footnotesize{\multirow{2}{*}{SALICON}} & \footnotesize{Soft Attention}  & 69.0 & 50.9 & 36.1 &	25.4 & 22.5 & 49.9 &	70.8 \\ \
    & \footnotesize{\textbf{Saliency+Context Attention}} & \textbf{69.2} & \textbf{51.4} & \textbf{37.2} & \textbf{26.9} & \textbf{22.9} & \textbf{50.4} & \textbf{73.3}  \\ \hline 
    \footnotesize{\multirow{2}{*}{COCO}} & \footnotesize{Soft Attention}  & 70.6 & 53.0 & 38.3 & 27.5 & 24.3 & 51.8 & 87.9 \\
    & \footnotesize{\textbf{Saliency+Context Attention}} & \textbf{70.8} & \textbf{53.6}  & \textbf{39.1} & \textbf{28.4} & \textbf{24.8} & \textbf{52.1} & \textbf{89.8}   \\ \hline 
    \footnotesize{Flickr8k} &  \footnotesize{Soft Attention}  & 59.9 & 41.8 & 27.9 & 18.2 & 19.8 & 45.0 & 47.7 \\
    \footnotesize{(Validation)} & \footnotesize{\textbf{Saliency+Context Attention}} & \textbf{62.8} & \textbf{44.5}  & \textbf{30.2} & \textbf{19.9} & \textbf{20.3} & \textbf{46.5} & \textbf{50.1}  \\ \hline 
    \footnotesize{Flickr8k} & \footnotesize{Soft Attention}  & 61.0 & 43.2 & 29.6 &	20.1 & 20.8 & 46.5 & 53.2 \\ 
    \footnotesize{(Test)}  & \footnotesize{\textbf{Saliency+Context Attention}} & \textbf{63.5} & \textbf{45.6}  & \textbf{31.5} & \textbf{21.2} &  \textbf{21.1}  & \textbf{47.5} & \textbf{54.1}   \\ \hline
    \footnotesize{Flickr30k} & \footnotesize{Soft Attention} & \textbf{61.9} &	\textbf{43.3}  & 29.7 & 20.2 & 19.9 & 44.8 & 43.2    \\
    \footnotesize{(Validation)} & \footnotesize{\textbf{Saliency+Context Attention}}  & 61.3  & \textbf{43.3}  & \textbf{30.1} & \textbf{20.9} & \textbf{20.2} &	\textbf{45.0} & \textbf{44.5} \\ \hline
    \footnotesize{Flickr30k} & \footnotesize{Soft Attention} & \textbf{61.9} & 43.4 & 29.9 & 20.5 &	19.8 & 44.5 & 44.2 \\ 
    \footnotesize{(Test)}  & \footnotesize{\textbf{Saliency+Context Attention}} & 61.5 &	\textbf{43.8}  & \textbf{30.5} & \textbf{21.3} &  \textbf{20.0}  & \textbf{45.2} & \textbf{46.4}  \\ \hline
    \footnotesize{\multirow{2}{*}{PASCAL-50S}} & \footnotesize{Soft Attention}  & \textbf{82.4} & 70.0 & 57.0 & 45.1 & 32.8 & 65.9 & 70.4 \\
    & \footnotesize{\textbf{Saliency+Context Attention}} & \textbf{82.4} & \textbf{70.2} & \textbf{57.5} & \textbf{45.7} & \textbf{32.9} & \textbf{66.3} & \textbf{70.7} \\ \hline 
    \end{tabular}
    \end{small}
    \end{center}
\end{table}

As mentioned, we perform experiments over five different datasets. For the SALICON dataset, since its images have all the same size of $480\times640$, we keep the original size of these images, thus obtaining $L=15\times20=300$. For all other datasets, which are composed of images with different sizes, we set the input size to $480\times480$ obtaining $L=15\times15=225$. Since saliency maps are exploited inside the proposed saliency-context attention model, we resize the SALICON saliency maps to have a size of $15\times20$ while, for all other datasets, we resize them to $15\times15$.

All experiments are performed by using the Adam optimizer~\cite{kingma2014adam} with Nestorov momentum~\cite{sutskever2013importance} using an initial learning rate of $0.001$ and batch size $64$. The hidden state dimension is set to $1024$ while the embedding size to $512$. For all datasets, we choose a vocabulary size equal to the number of words which appear at least 5 times in training and validation captions.

\subsection{Quantitative results and comparisons with baselines}
To assess the performance of our method, and to investigate the hypotheses behind it, we first compare with the classic Soft Attention approach, and we then build three baselines in which saliency is used to condition the generative process.

\textbf{Soft Attention~\cite{icml2015xuc15}}:~The visual input to the LSTM is computed via the Soft Attention mechanism to attend at different locations of the image, without considering salient and non-salient regions. A single feed forward network is in charge of producing attention values, which can be obtained by replacing Eq.~\ref{eq:att2} with
\begin{equation}
e_{ti} = v^T_e \cdot \phi(W_{ae}\cdot\mathbf{a}_i + W_{he} \cdot \mathbf{h}_{t-1}).
\end{equation}
This approach is equivalent to the one proposed in~\cite{icml2015xuc15}, although some implementation details are different. In order to achieve a fair evaluation, we use activations from the ResNet-50 model instead of the VGG-19, and we do not include the doubly stochastic regularization trick. For this reason, the numerical results that we report are not directly comparable with those in the original paper (ours are in general higher than the original ones).

\textbf{Saliency pooling}:~Visual features from the CNN are multiplied at each location by the corresponding saliency value, and then summed, without any attention mechanism. In this case the visual input of the LSTM is not time dependent, and salient regions are given more focus than non-salient ones. Comparing with Eq.~\ref{eq:att_sum}, it can be seen as a variation of the Soft Attention in which the network always focuses on salient regions.
\begin{equation}
\hat{\mathbf{v}_t} = \hat{\mathbf{v}} = \sum_{i=1}^L s_i \mathbf{a}_i
\end{equation}

\begin{table}[t]
    \begin{center}
    \renewcommand{\arraystretch}{1.2}
    \caption{Comparison with image captioning with saliency baselines. While the use of machine attention strategies is beneficial (see \texttt{Saliency pooling} vs. \texttt{Attention on Saliency}), saliency and context are both important for captioning. The use of different attention paths for saliency and context also enhances the performance (see \texttt{Saliency+Context Attention (with weight sharing)} vs. \texttt{Saliency+Context Attention}).}
    \label{tab:captioning_3}
    \begin{small}
    \begin{tabular}{|c|c|ccccccc|}
    \hline
    \footnotesize{Dataset} 	& \footnotesize{Model} & \footnotesize{B@1} & \footnotesize{B@2} &  \footnotesize{B@3} & \footnotesize{B@4} & \footnotesize{METEOR} &  \footnotesize{ROUGE$_L$} &  \footnotesize{CIDEr} \\ \hline \hline 
    \footnotesize{\multirow{4}{*}{SALICON}} 
    &  \footnotesize{Saliency pooling}  & 66.1 & 47.8 &	33.7 &	24.0 &	21.1 &	47.9 &	62.4 \\ 
    &  \footnotesize{Attention on saliency}  & 68.8 & 51.3 & 37.0 &	26.5 &	22.7 &	50.1 &	71.3 \\ 
    &   \footnotesize{Saliency+Cont. Att. (weight sh.)}  & 68.9 &	51.3 &	36.8 & 26.3 & 22.6 & 50.2 & 71.4 \\ 
    & \footnotesize{Saliency+Context Attention} & \textbf{69.2} & \textbf{51.4} & \textbf{37.2} & \textbf{26.9} & \textbf{22.9} & \textbf{50.4} & \textbf{73.3}  \\ \hline 
    \footnotesize{\multirow{4}{*}{COCO}} 
    & \footnotesize{Saliency pooling}  & 68.6 &	50.9 & 36.3 & 25.8 & 23.3 &	50.2 & 81.4 \\ 
    &  \footnotesize{Attention on saliency} & 70.4 & 53.2 & 38.6 & 27.6 & 24.1 & 51.6 &	86.6 \\ 
    &   \footnotesize{Saliency+Cont. Att. (weight sh.)}  & 70.4 & 53.1  & 38.8 & 28.2 & 24.7 & \textbf{52.1} & 89.4  \\ 
    & \footnotesize{Saliency+Context Attention} & \textbf{70.8} & \textbf{53.6}  & \textbf{39.1} & \textbf{28.4} & \textbf{24.8} & \textbf{52.1} & \textbf{89.8}  \\ \hline 
     &  \footnotesize{Saliency pooling}  & 56.1 & 37.7 & 24.3 &	15.6 & 18.3 & 42.8 & 37.0 \\ 
    \footnotesize{Flickr8k}  &  \footnotesize{Attention on saliency}  & 58.7 & 40.4 & 26.8 & 17.6 &	19.7 & 45.1 & 44.7 \\ 
     \footnotesize{(Validation)} &  \footnotesize{Saliency+Cont. Att. (weight sh.)}  & 62.0 & 43.9 & 29.6 &	19.8 & 20.2 &	45.7 &	\textbf{50.2} \\ 
     & \footnotesize{Saliency+Context Attention} & \textbf{62.8} & \textbf{44.5}  & \textbf{30.2} & \textbf{19.9} & \textbf{20.3} & \textbf{46.5} & 50.1  \\ \hline 
     &  \footnotesize{Saliency pooling}  & 56.5 & 37.8 & 24.6 &	16.2 &	18.5 &	42.9 &	37.7 \\ 
   \footnotesize{Flickr8k} &  \footnotesize{Attention on saliency}  & 59.6 & 42.2 & 28.7 & 19.5 & 20.7 & 46.1 & 50.1 \\
    \footnotesize{(Test)} &  \footnotesize{Saliency+Cont. Att. (weight sh.)}  & 62.4 & 44.2 & 29.9 & 19.7 &	21.1 & 46.7 & 51.7 \\ 
      & \footnotesize{Saliency+Context Attention} & \textbf{63.5} &	\textbf{45.6}  & \textbf{31.5} & \textbf{21.2} &  \textbf{21.1}  & \textbf{47.5} & \textbf{54.1}  \\ \hline
     &  \footnotesize{Saliency pooling}  & 58.7 & 40.5 & 27.1 &	18.4 & 18.3 & 43.0 & 34.2 \\  
     \footnotesize{Flickr30k} &  \footnotesize{Attention on saliency} & \textbf{63.0} & \textbf{44.5} & \textbf{30.8} & \textbf{21.3} & 19.4 & 44.7 & 43.5 \\ 
     \footnotesize{(Validation)} &  \footnotesize{Saliency+Cont. Att. (weight sh.)}  & 62.0 & 43.8 & 30.0 & 20.5 & 19.7 & 44.6 & 43.3 \\ 
     & \footnotesize{Saliency+Context Attention} & 61.3  & 43.3  & 30.1  & 20.9 & \textbf{20.2} & \textbf{45.0} & \textbf{44.5} \\ \hline 
    &  \footnotesize{Saliency pooling}  & 58.3 & 40.6 & 27.5 & 18.6 & 18.7 & 43.0 &	36.2 \\ 
    \footnotesize{Flickr30k}  &  \footnotesize{Attention on saliency}  & \textbf{62.5} & \textbf{44.2} & \textbf{30.5} & 21.0 & 19.6 & 44.9 &	45.0 \\ 
    \footnotesize{(Test)} &  \footnotesize{Saliency+Cont. Att. (weight sh.)} & 61.7 & 43.7 & 30.0 & 20.4 & 19.6 & 44.2 & 43.1 \\ 
      & \footnotesize{Saliency+Context Attention} & 61.5 &	43.8  & \textbf{30.5} & \textbf{21.3} & \textbf{20.0}  & \textbf{45.2} & \textbf{46.4} \\ \hline
      \footnotesize{\multirow{4}{*}{PASCAL-50S}} 
    & \footnotesize{Saliency pooling} & 79.9 & 67.1 & 53.6 & 41.8 & 31.4 & 64.1 & 65.3 \\ 
    &  \footnotesize{Attention on saliency} & \textbf{82.4} & \textbf{70.3} & 57.4 & 45.5 & 32.7 & \textbf{66.3} &	70.2 \\ 
    &   \footnotesize{Saliency+Cont. Att. (weight sh.)} & 82.0 & 69.7 & 56.4 &	44.2 & 32.7 & 65.2 & 70.0 \\ 
    & \footnotesize{Saliency+Context Attention} & \textbf{82.4} & 70.2 & \textbf{57.5} & \textbf{45.7} & \textbf{32.9} & \textbf{66.3} & \textbf{70.7}  \\  \hline 
    \end{tabular}
    \end{small}
    \end{center}
\end{table}

\textbf{Attention on saliency}:~This is an extension of the Soft Attention approach in which saliency is used to modulate attention values at each location. The attention mechanism, therefore, is conditioned to attend salient regions with higher probability, and to ignore non-salient regions.
\begin{equation}
e_{ti} = s_i \cdot v^T_e \cdot \phi(W_{ae}\cdot\mathbf{a}_i + W_{he} \cdot \mathbf{h}_{t-1})
\end{equation}

\textbf{Attention on saliency and context (with weight sharing)}:~The attention mechanism is aware of salient and context regions, but weights used to compute the attentive scores of salient and context are shared, excluding the $v_e^T$ vectors. Notice that, if those were shared too, this baseline would be equivalent to the Soft Attention one.
\begin{align}
    e_{ti} &= s_i \cdot e_{ti}^{sal} + (1-s_i) \cdot e_{ti}^{ctx} \\
    e_{ti}^{sal} &= v_{e, sal}^T \cdot \phi (W_{ae} \cdot \mathbf{a}_i + W_{he} \cdot \mathbf{h}_{t-1}) \\
    e_{ti}^{ctx} &= v_{e, ctx}^T \cdot \phi (W_{ae} \cdot \mathbf{a}_i + W_{he} \cdot \mathbf{h}_{t-1}) 
\end{align}

It is straightforward also to notice that our proposed approach is equivalent to the last baseline, without weight sharing.

In Table~\ref{tab:captioning} we first compare the performance of our method with respect to the Soft Attention approach, to assess the superior performance of the proposal with respect to the published state of the art. We report results on all the datasets, both on validation and test sets, with respect to all the automatic metrics described in Section~\ref{sec:dataset_metrics}. As it can be seen, the proposed approach always overcomes by a significant margin the Soft Attention approach, thus experimentally confirming the benefit of having two separate attention paths, one for salient and one for non-salient regions, and the role of saliency as a conditioning for captioning. In particular, on the METEOR metric, the relative improvement ranges from $\frac{32.9 - 32.8}{32.8} = 0.30\%$ on the PASCAL-50S to $\frac{20.3 - 19.8}{19.8} = 2.53\%$ of the Flickr8k validation set.

In Table~\ref{tab:captioning_3}, instead, we compare our approach with the three baselines that incorporate saliency. Firstly, it can be observed that the Saliency pooling baseline usually performs worse than the Soft Attention, thus demonstrating that always attending to salient locations is not sufficient to achieve good captioning results. When plugging in attention, like in the Saliency Attention baseline, numerical results are a bit higher, thanks to a time-dependent attention, but still far from the performance achieved by the complete model. It can also be noticed that, even though this baseline does not take into account the context, it sometimes achieves better results than the Soft Attention model (such as in the case of SALICON, with respect to the METEOR metric). Finally, we notice that the baseline with attention on saliency and context, and with weight sharing, is better than Saliency Attention, further confirming the benefit of including the context. Having two completely separated attention paths, such as in our model, is anyway important, as demonstrated by the numerical results of this last baseline with respect to those of our method.

\begin{table}[tb]
    \begin{center}
    \renewcommand{\arraystretch}{1.2}
    \caption{Comparison with existing saliency-boosted captioning models.}
    \label{tab:comparison}
    \begin{small}
    \begin{tabular}{|c|c|cccc|}
    \hline
    \footnotesize{Dataset} 	& \footnotesize{Model} & \footnotesize{B@4} & \footnotesize{METEOR} &  \footnotesize{ROUGE$_L$} &  \footnotesize{CIDEr} \\ \hline \hline 
    \footnotesize{\multirow{2}{*}{SALICON}} 
    & \footnotesize{Sugano~\etal\cite{sugano2016seeing}} & 24.5 & 21.9 & \textbf{52.4} & 63.8 \\ 
    & \footnotesize{\textbf{Ours}} & \textbf{26.9} & \textbf{22.9} & 50.4 & \textbf{73.3}  \\ \hline 
    \footnotesize{\multirow{3}{*}{COCO}} 
    & \footnotesize{Tavakoli~\etal\cite{tavakoli2017can} (GBVS)} & \textbf{28.7} & 23.5 & 51.2 & 84.1 \\ 
    & \footnotesize{Tavakoli~\etal\cite{tavakoli2017can} (iSEEL)}  & 28.3 & 23.5 & 50.8 & 83.6 \\ 
    & \footnotesize{\textbf{Ours}} & 28.4 & \textbf{24.8} & \textbf{52.1} & \textbf{89.8}  \\ \hline 
    \footnotesize{\multirow{2}{*}{Flickr30k (Test)}} 
    & \footnotesize{Ramanishka~\etal\cite{ramanishka2016top}} & -- & 18.3 & -- & -- \\ 
    & \footnotesize{\textbf{Ours}} & 21.3 & \textbf{20.0} & 45.2 & 46.4  \\ \hline 
    \footnotesize{\multirow{3}{*}{PASCAL-50S}} 
    & \footnotesize{Tavakoli~\etal\cite{tavakoli2017can} (GBVS)}  & 40.0  & 30.2 & 63.5 & 61.5 \\ 
    & \footnotesize{Tavakoli~\etal\cite{tavakoli2017can} (iSEEL)}  & 39.6 & 30.2  & 63.2 & 61.4 \\ 
    & \footnotesize{\textbf{Ours}} & \textbf{45.7} & \textbf{32.9} & \textbf{66.3} & \textbf{70.7}   \\ \hline
    \end{tabular}
    \end{small}
    \end{center}
\end{table}

\subsection{Comparisons with other saliency-boosted captioning models}
We also compare to existing captioning models that incorporate saliency during the generation of image descriptions. In particular, we compare to the model proposed in~\cite{sugano2016seeing}, which exploited human fixation points, to the work by Tavakoli~\etal\cite{tavakoli2017can} which reports experiments on Microsoft COCO and on PASCAL-50S, and to the proposal by Ramanishka~\etal\cite{ramanishka2016top} which used convolutional activations as a proxy for saliency.

Table~\ref{tab:comparison} shows the results on the three considered datasets in term of BLEU@4, METEOR, ROUGE$_L$ and CIDEr.
We compare our solutions to both versions of the model presented in~\cite{tavakoli2017can}. The GBVS version exploits saliency maps calculated by using a traditional bottom-up model~\cite{harel2006graph}, while the other one includes saliency maps extracted from a deep convolutional network~\cite{tavakoli2017exploiting}.

Overall, results show that the proposed Saliency and Context Attention model can overcome the other methods on different metrics, thus confirming the strategy of including two attention paths. In particular, on the METEOR metric, we obtain a relative improvement of $4.57\%$ on the SALICON dataset, $5.53\%$ on the Microsoft COCO and $8.94\%$ on the PASCAL-50S.

\subsection{Analysis of generated captions}
\begin{table}[tb]
    \begin{center}
    \renewcommand{\arraystretch}{1.2}
    \caption{Statistics on vocabulary size and diversity of the generated captions. Including saliency and context in two different machine attention paths (\texttt{Saliency+Context attention}) produces different captions with respect to the traditional machine attention approach (\texttt{Soft Attention}), while preserving almost the same diversity statistics.}
    \label{tab:captioning_2}
    \begin{small}
    \begin{tabular}{|c|c|ccccc|}
    \hline
    \footnotesize{Dataset} 	& \footnotesize{Model} & \footnotesize{Div-1} & \footnotesize{Div-2} &  \footnotesize{Vocabulary} & \footnotesize{\% novel sent.} & \footnotesize{\% different sent.} \\ \hline \hline 
    \footnotesize{\multirow{2}{*}{SALICON}} & \footnotesize{Soft Attention}  & 0.0136 &	0.0498 & 658 & 95.22\% & \multirow{2}{*}{95.34\%} \\ \
    & \footnotesize{\textbf{Saliency+Context Attention}} & 0.0125 & 0.0549 & 614 & 93.12\% & \\ \hline 
    \footnotesize{\multirow{2}{*}{COCO}} & \footnotesize{Soft Attention}  & 0.0038 & 0.0187 & 1490 & 81.81\% & \multirow{2}{*}{93.80\%} \\
    & \footnotesize{\textbf{Saliency+Context Attention}} & 0.0037 & 0.0182 & 1444 & 78.02\% & \\ \hline  
    \footnotesize{Flickr8k} &  \footnotesize{Soft Attention} & 0.0367 & 0.1026 & 389 & 98.30\% & \multirow{2}{*}{97.90\%} \\
    \footnotesize{(Validation)} & \footnotesize{\textbf{Saliency+Context Attention}} & 0.0400 & 0.1094 & 411 & 99.30\% & \\ \hline 
    \footnotesize{Flickr8k} & \footnotesize{Soft Attention} & 0.0385 & 0.1041 & 404 & 98.50\% & \multirow{2}{*}{97.60\%} \\ 
    \footnotesize{(Test)}  & \footnotesize{\textbf{Saliency+Context Attention}} & 0.0419 & 0.1119 & 423 & 99.60\% & \\ \hline 
    \footnotesize{Flickr30k} & \footnotesize{Soft Attention} & 0.0577 & 0.1445 & 699 & 99.90\% &  \multirow{2}{*}{98.62\%} \\
    \footnotesize{(Validation)} & \footnotesize{\textbf{Saliency+Context Attention}}& 0.0565 & 0.1439 & 715  & 99.61\% & \\ \hline  
    \footnotesize{Flickr30k} & \footnotesize{Soft Attention} & 0.0580 & 0.1508 & 682 & 99.90\% & \multirow{2}{*}{98.20\%} \\
    \footnotesize{(Test)}  & \footnotesize{\textbf{Saliency+Context Attention}} & 0.0585 & 0.1549 & 711 & 99.70\% & \\ \hline  
    \footnotesize{\multirow{2}{*}{PASCAL-50S}} & \footnotesize{Soft Attention} & 0.0475 & 0.1379 & 465 & 97.10\% & \multirow{2}{*}{94.80\%} \\
    & \footnotesize{\textbf{Saliency+Context Attention}} & 0.0468 & 0.1359 & 456 & 96.40\% & \\ \hline
    \end{tabular}
    \end{small}
    \end{center}
\end{table}
We further collect statistics on captions generated by our method and by the Soft Attention model, to quantitatively assess the quality of generated captions. Firstly, we define three metrics which evaluate the vocabulary size and the difference between the corpus of captions generated by the two models and the ground-truth:
\begin{itemize}
\item \textit{Vocabulary size}: number of unique words generated in all captions;
\item \textit{Percentage of novel sentences}: percentage of generated sentences which are not seen in the training set;
\item \textit{Percentage of different sentences}: percentage of images which are described differently by the two models;
\end{itemize}
Then, we measure the diversity of the set of captions generated by each of the two models, via the following two metrics~\cite{shetty2017speaking}:
\begin{itemize}
\item \textit{Div-1}: ratio of number of unique unigrams in a set of captions to the number of words in the same set. Higher is more diverse.
\item \textit{Div-2}: ratio of number of unique bigrams in a set of captions to the number of words in the same set. Higher is more diverse.
\end{itemize}

In Table~\ref{tab:captioning_2} we compare the set of captions generated by our model with that generated by the Soft Attention baseline. Although our model features a slight reduction of the vocabulary size on SALICON, COCO and PASCAL-50S, captions generated by the two models are very often different, thus confirming that the two approaches have learned different captioning models. Moreover, the diversity and the number of novel sentences of the Soft Attention approach are entirely preserved.

\subsection{Analysis of attentive states}
\begin{figure}[t]
\begin{center}
    \begin{minipage}[t]{0.48\linewidth}
    \includegraphics[height=.30\textwidth]{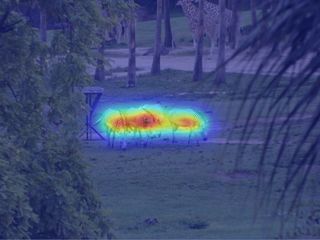}  
    \includegraphics[height=.30\textwidth]{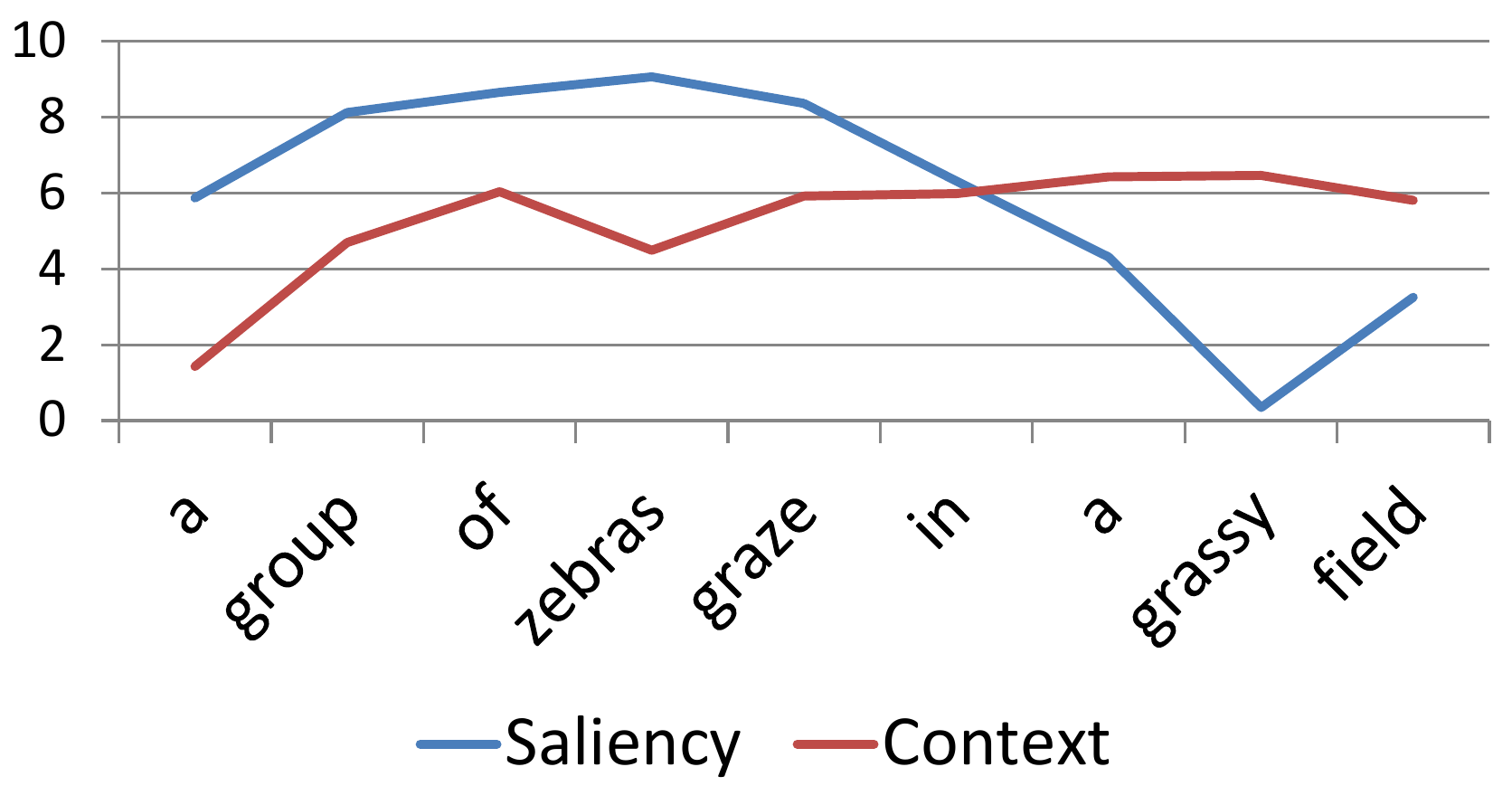}  
    \end{minipage}
    \begin{minipage}[t]{0.48\linewidth}
    \includegraphics[height=.30\textwidth]{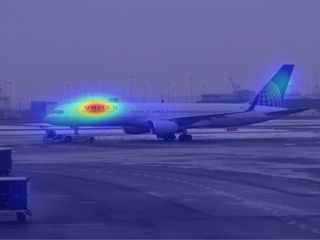}  
    \includegraphics[height=.30\textwidth]{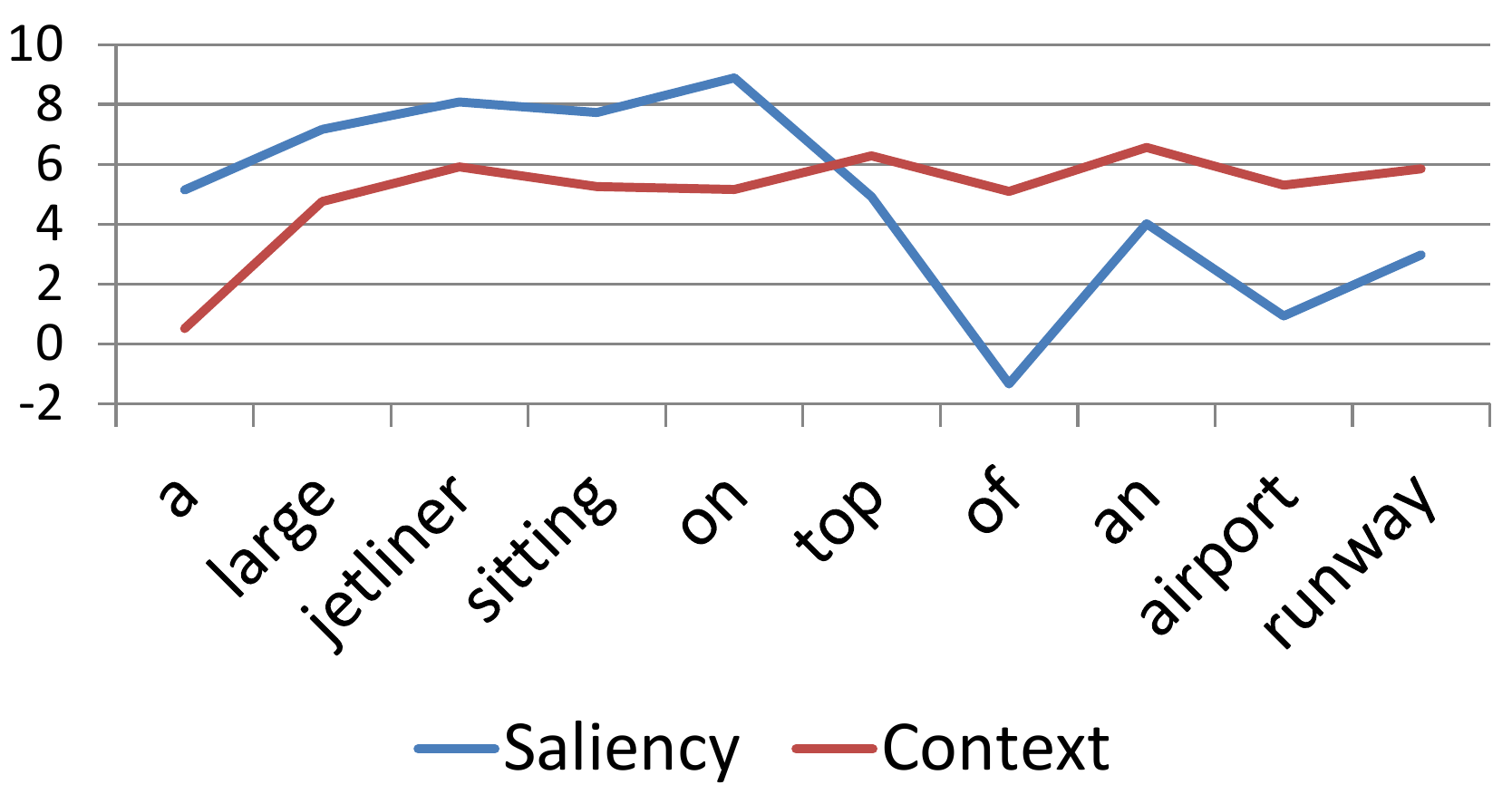}  
    \end{minipage}
    \begin{minipage}[t]{0.48\linewidth}
    \includegraphics[height=.30\textwidth]{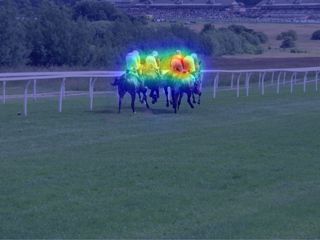}  
    \includegraphics[height=.30\textwidth]{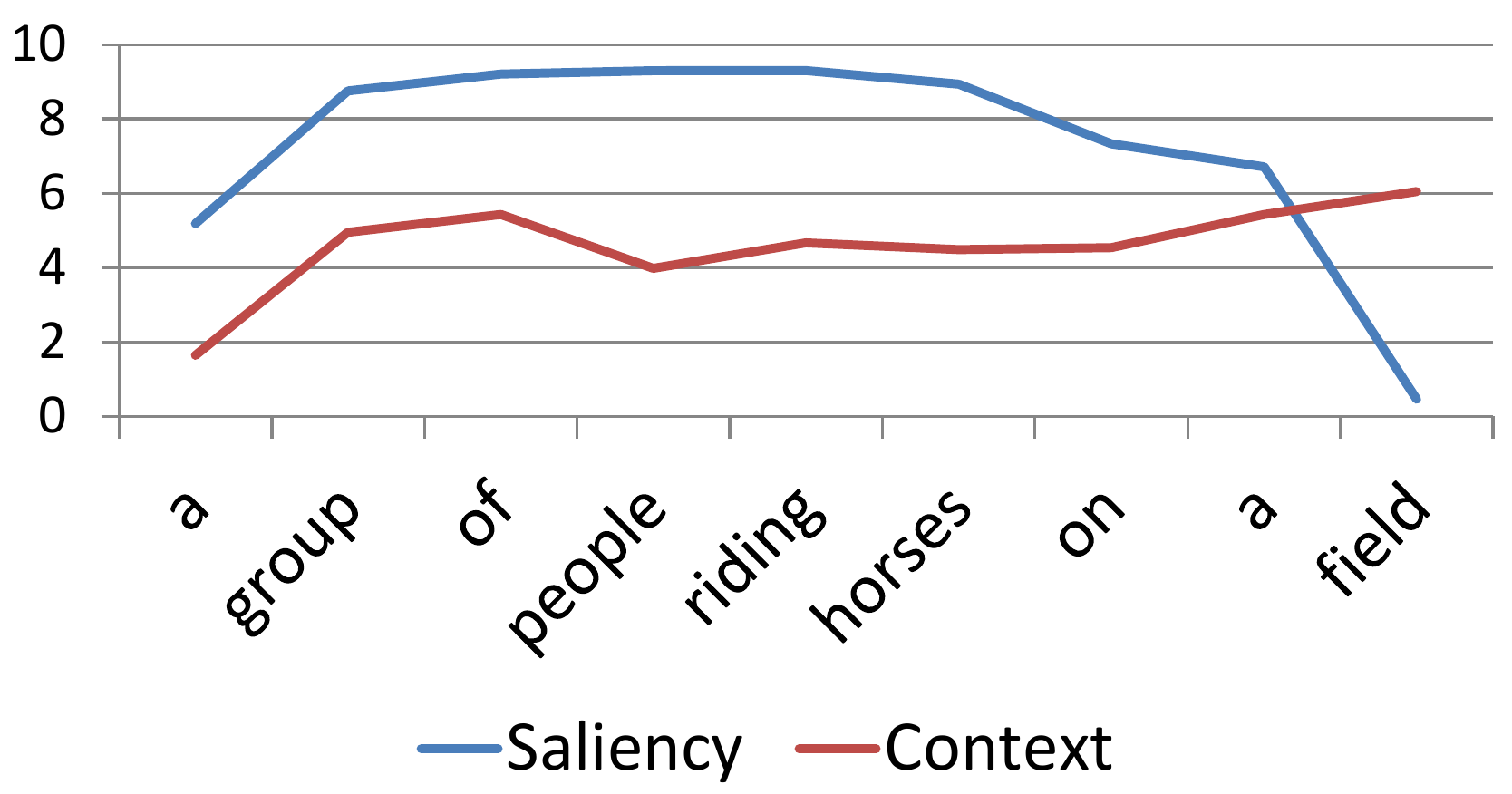}  
    \end{minipage}
    \begin{minipage}[t]{0.48\linewidth}
    \includegraphics[height=.30\textwidth]{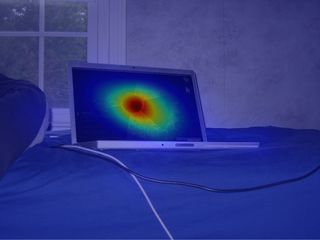}  
    \includegraphics[height=.30\textwidth]{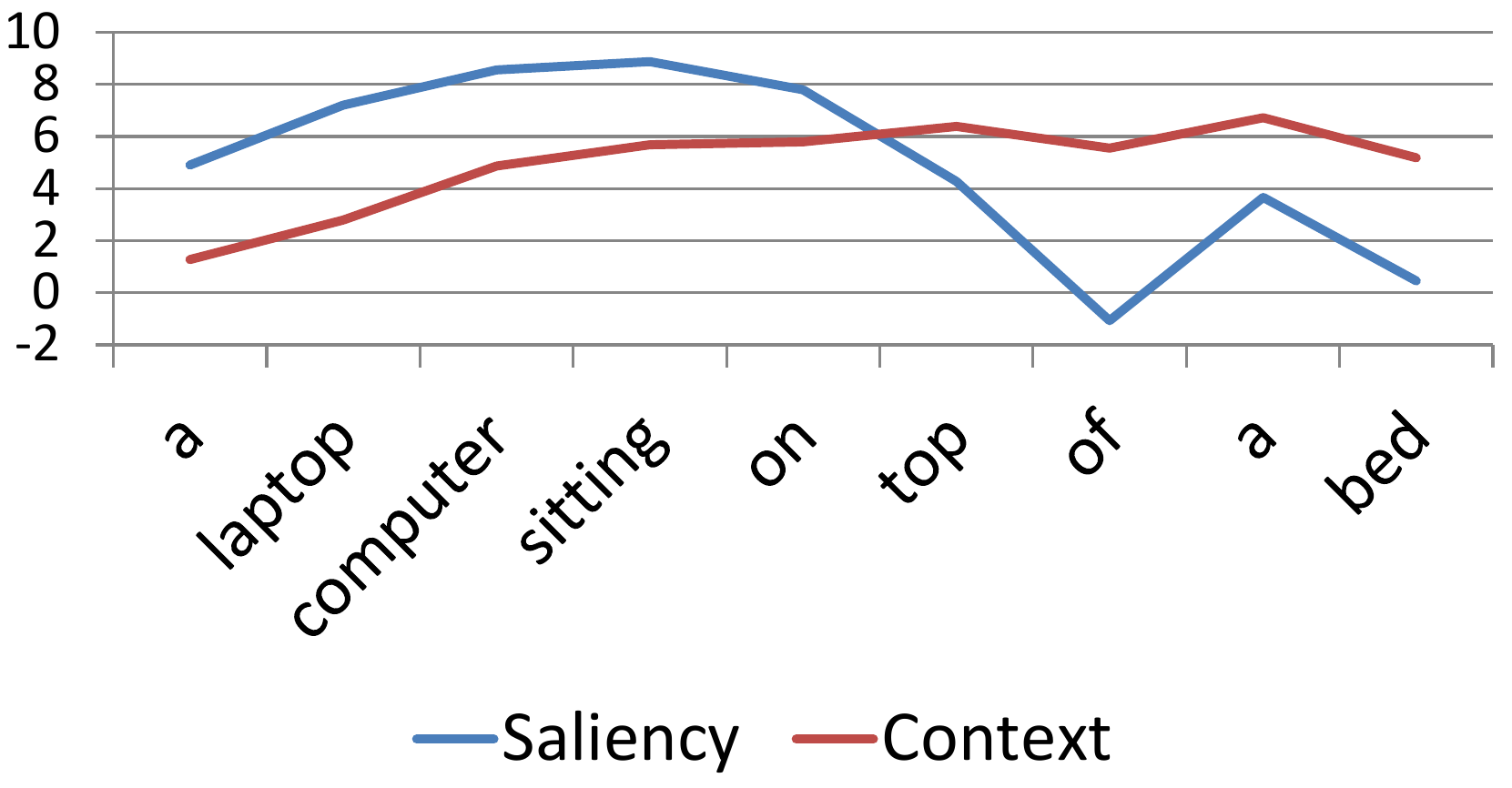}  
    \end{minipage}
    \begin{minipage}[t]{0.48\linewidth}
    \includegraphics[height=.30\textwidth]{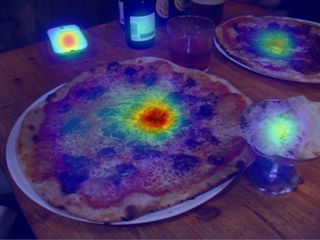}  
    \includegraphics[height=.30\textwidth]{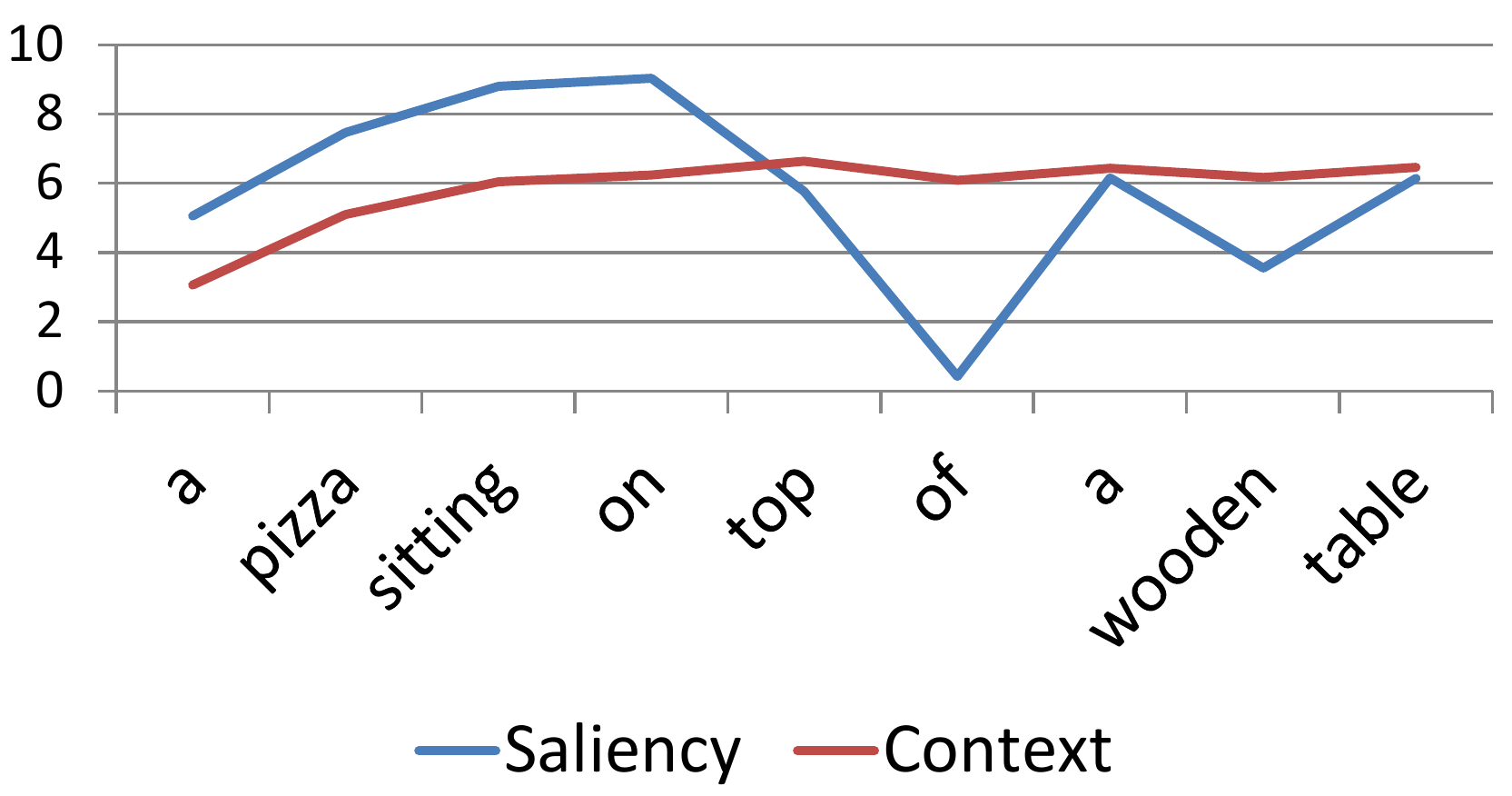}  
    \end{minipage}
    \begin{minipage}[t]{0.48\linewidth}
    \includegraphics[height=.30\textwidth]{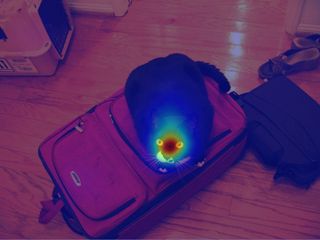}  
    \includegraphics[height=.30\textwidth]{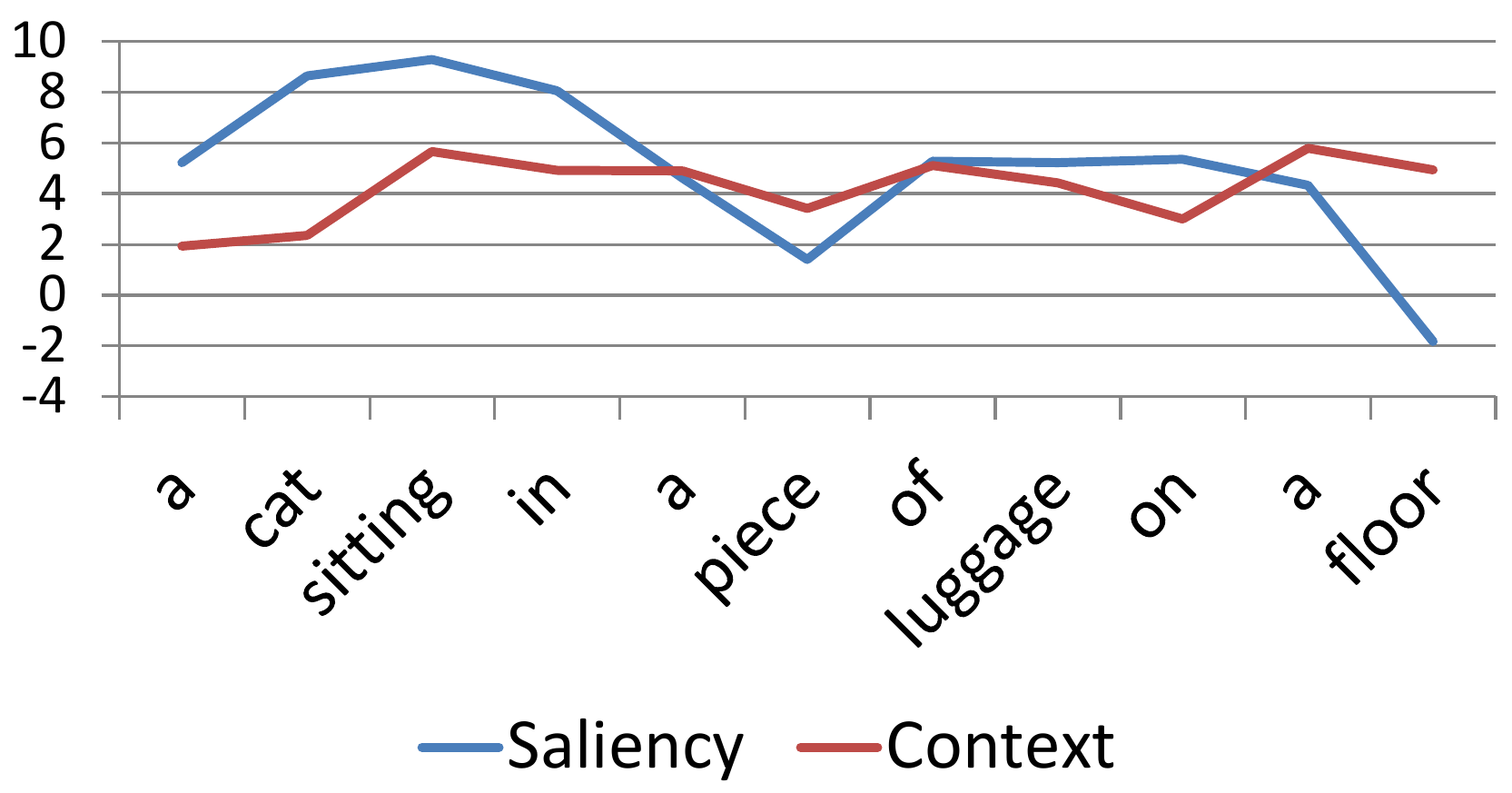}  
    \end{minipage}
    \begin{minipage}[t]{0.48\linewidth}
    \includegraphics[height=.30\textwidth]{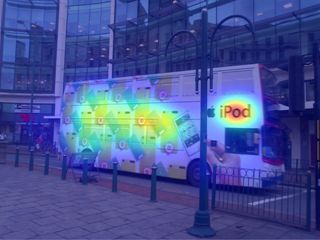}  
    \includegraphics[height=.30\textwidth]{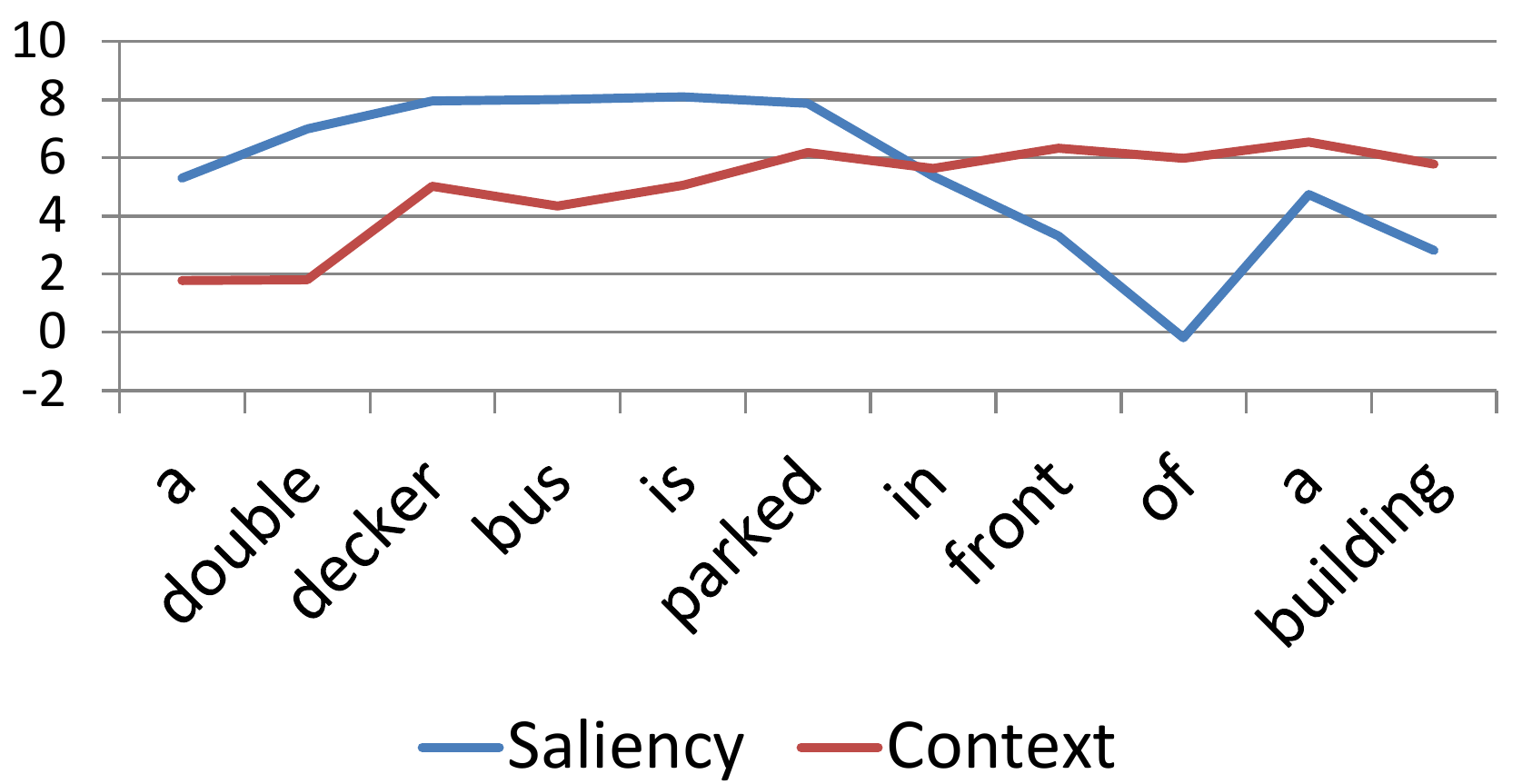}  
    \end{minipage}
    \begin{minipage}[t]{0.48\linewidth}
    \includegraphics[height=.30\textwidth]{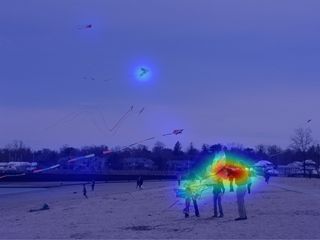}  
    \includegraphics[height=.30\textwidth]{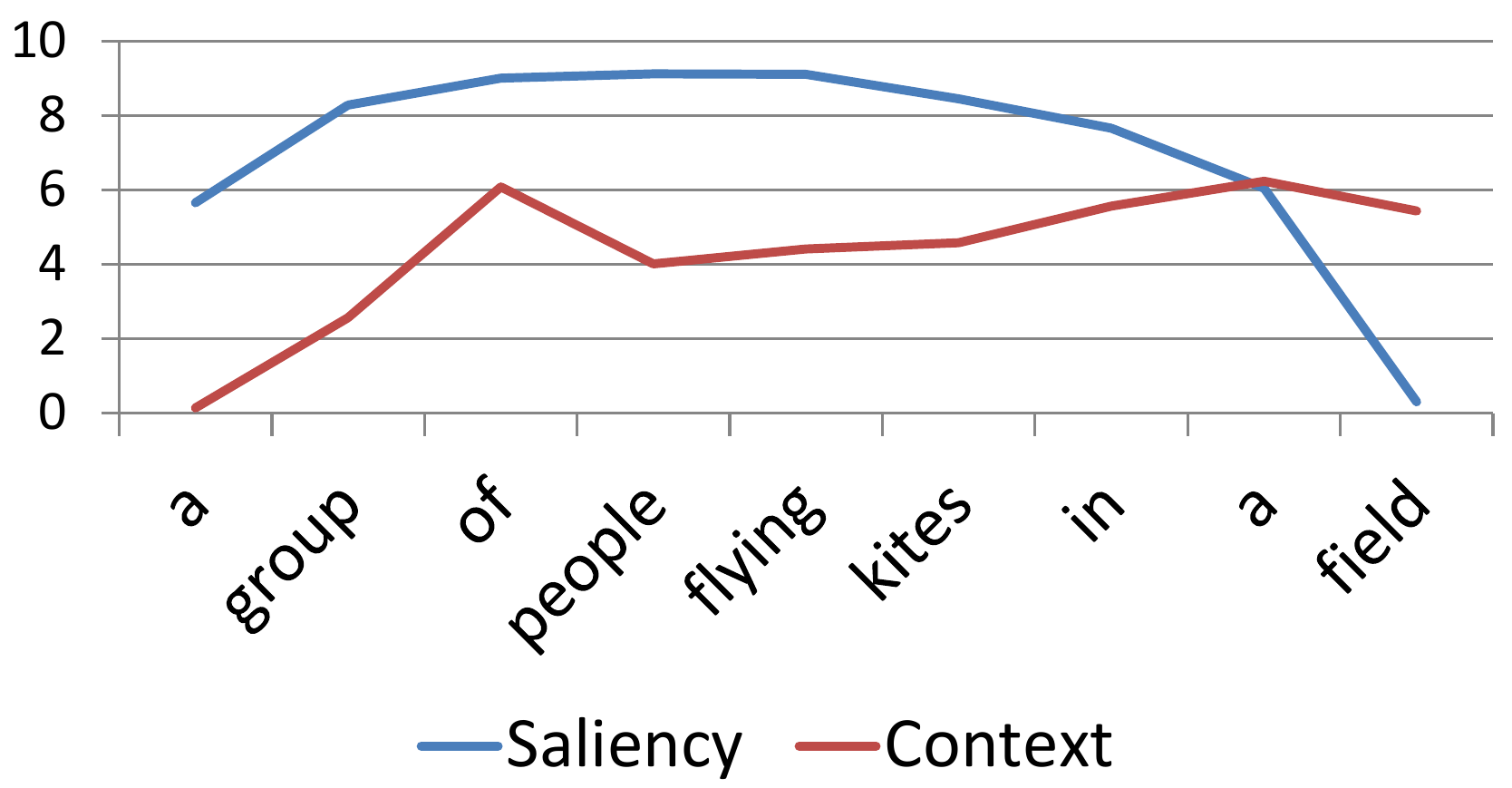}  
    \end{minipage}
\end{center}
\caption{Examples of attention weights changes between saliency and context along with the generation of captions (best seen in color). Images are from the Microsoft COCO dataset~\cite{lin2014microsoft}.}
\label{fig:attention}
\end{figure}
The selection of a location in our model is based on the competition between the saliency attentive path and the context attentive path (see Eq.~\ref{eq:att2}). To investigate how the two paths interact and contribute to the generation of a word, in Figure~\ref{fig:attention} we report, for several images from the Microsoft COCO dataset, the changes in attention weights between the two paths. Specifically, for each image we report the average of $e_{ti}^{sal}$ and $e_{ti}^{ctx}$ values at each timestep, along with a visualization of its saliency map. It is interesting to see how the model was able to correctly exploit the two attention paths for generating different parts of the caption, and how generated words correspond in most cases to the attended regions. For example, in the case of the first image (``a group of zebras graze in a grassy field''), the saliency attentive path is more active than the context path during the generation of words corresponding to the ``group of zebras'', which are captured by saliency. Instead, when the model has to describe the context (``in a grassy field''), the saliency attentive path has lower weights with respect to the context attentive path. The same can be observed for all the reported images; it can also be noticed that generated captions tend to describe both salient objects and the context, and that usually the salient part, which is also the most important, is described before the context.

\begin{figure}[t]
\begin{center}
    \begin{minipage}[t]{0.32\linewidth}
    \includegraphics[width=.49\textwidth]{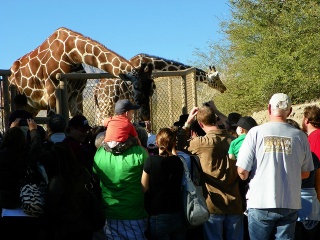}  
    \includegraphics[width=.49\textwidth]{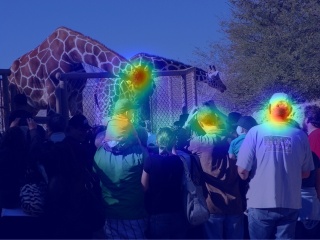}  
    \scriptsize{
        \textbf{Ground-truth}: A group of people that are standing around giraffes. \\
        \textbf{Saliency+Context Attention}: A group of people standing around a giraffe. \\
        \textbf{Without saliency}: A group of people standing around a stage with a group of people. \\
    }
    \end{minipage}
    \hspace{0.02cm}
    \begin{minipage}[t]{0.32\linewidth}
    \includegraphics[width=.49\textwidth]{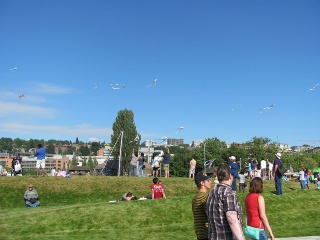}  
    \includegraphics[width=.49\textwidth]{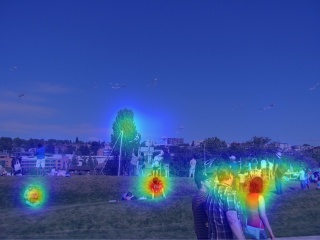}  
    \scriptsize{
        \textbf{Ground-truth}: A group of people at the park with some flying kites. \\
        \textbf{Saliency+Context Attention}: A group of people flying kites in a park. \\
        \textbf{Without saliency}: A group of people standing on top of a lush green field. \\
    }
    \end{minipage}
    \hspace{0.02cm}
    \begin{minipage}[t]{0.32\linewidth}
    \includegraphics[width=.49\textwidth]{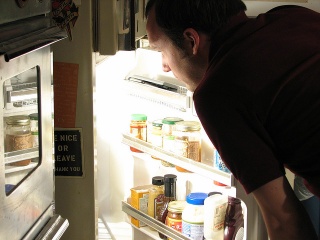} 
    \includegraphics[width=.49\textwidth]{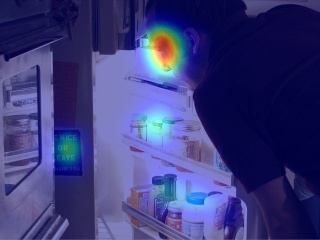}  
    \scriptsize{
        \textbf{Ground-truth}: A man is looking into a home refrigerator. \\
        \textbf{Saliency+Context Attention}: A man is looking inside of a refrigerator. \\
        \textbf{Without saliency}: A man is making a refrigerator in a kitchen. \\
    }
    \end{minipage}
    \begin{minipage}[t]{0.32\linewidth}
    \includegraphics[width=.49\textwidth]{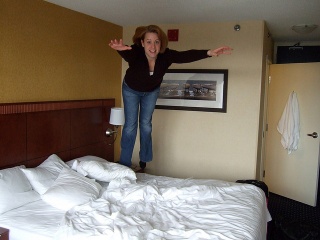} 
    \includegraphics[width=.49\textwidth]{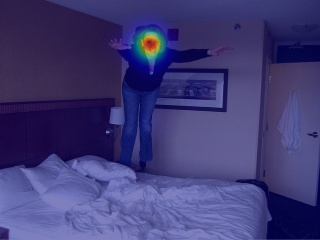}  
    \scriptsize{
        \textbf{Ground-truth}: A women who is jumping on the bed. \\
        \textbf{Saliency+Context Attention}: A woman is jumping up in a bed. \\
        \textbf{Without saliency}: A woman is playing with a remote control. \\
    }
    \end{minipage}
    \hspace{0.02cm}
    \begin{minipage}[t]{0.32\linewidth}
    \includegraphics[width=.49\textwidth]{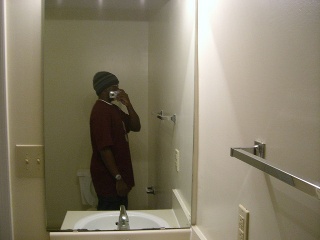}  
    \includegraphics[width=.49\textwidth]{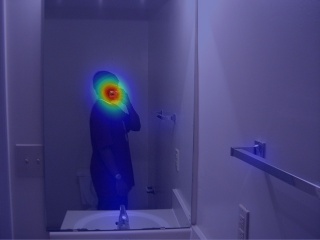}  
    \scriptsize{
        \textbf{Ground-truth}: A man takes a profile picture of himself in a bathroom mirror. \\
        \textbf{Saliency+Context Attention}: A person taking a picture of himself in a bathroom. \\
        \textbf{Without saliency}: A bathroom with a sink and a sink. \\
    }
    \end{minipage}
    \hspace{0.02cm}
    \begin{minipage}[t]{0.32\linewidth}
    \includegraphics[width=.49\textwidth]{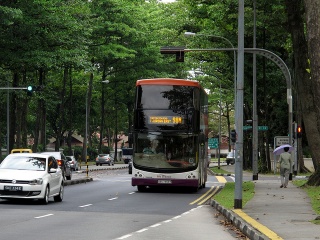}  
    \includegraphics[width=.49\textwidth]{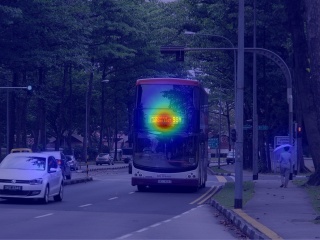}  
    \scriptsize{
        \textbf{Ground-truth}: A double decker bus driving down a street. \\
        \textbf{Saliency+Context Attention}: A double decker bus driving down a street. \\
        \textbf{Without saliency}: A bus is parked on the side of the road. \\
    }
    \end{minipage}
    \begin{minipage}[t]{0.32\linewidth}
    \includegraphics[width=.49\textwidth]{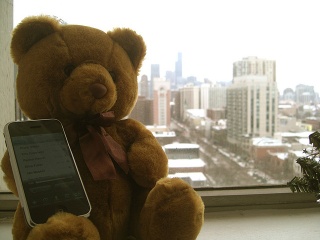} 
    \includegraphics[width=.49\textwidth]{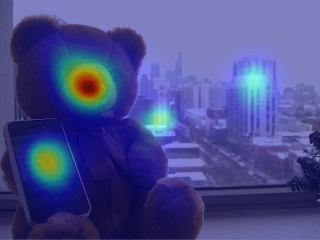}  
    \scriptsize{
        \textbf{Ground-truth}: A teddy bear holding a cell phone in front of a window with a view of the city. \\
        \textbf{Saliency+Context Attention}: A teddy bear sitting on a chair next to a window. \\
        \textbf{Without saliency}: A brown dog is sitting on a laptop keyboard. \\
    }
    \end{minipage}
    \hspace{0.02cm}
    \begin{minipage}[t]{0.32\linewidth}
    \includegraphics[width=.49\textwidth]{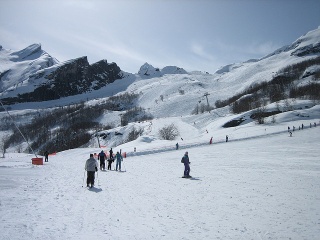} 
    \includegraphics[width=.49\textwidth]{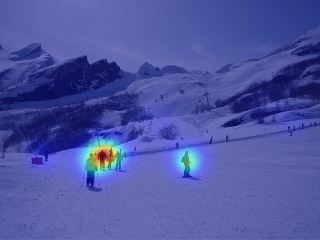}  
    \scriptsize{
        \textbf{Ground-truth}: A group of people riding down a snow covered slope. \\
        \textbf{Saliency+Context Attention}: A group of people riding skis down a snow covered slope. \\
        \textbf{Without saliency}: A group of people on skis in the snow. \\
    }
    \end{minipage}
    \hspace{0.02cm}
    \begin{minipage}[t]{0.32\linewidth}
    \includegraphics[width=.49\textwidth]{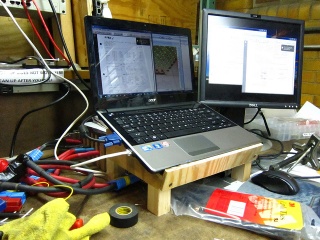} 
    \includegraphics[width=.49\textwidth]{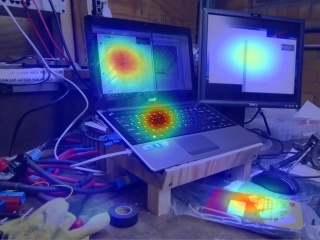}  
    \scriptsize{
        \textbf{Ground-truth}: A laptop computer sitting on top of a table. \\
        \textbf{Saliency+Context Attention}: A laptop computer sitting on a top of a desk. \\
        \textbf{Without saliency}: A desk with a laptop computer and a laptop. \\
    }
    \end{minipage}
    \begin{minipage}[t]{0.32\linewidth}
    \includegraphics[width=.49\textwidth]{images/examples/COCO_val2014_000000037751.jpg} 
    \includegraphics[width=.49\textwidth]{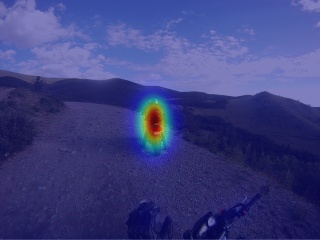}  
    \scriptsize{
        \textbf{Ground-truth}: A person on a motorcycle riding on a mountain. \\
        \textbf{Saliency+Context Attention}: A person riding a motorcycle on a road. \\
        \textbf{Without saliency}: A man on a bike with a bike in the background. \\
    }
    \end{minipage}
    \hspace{0.02cm}
    \begin{minipage}[t]{0.32\linewidth}
    \includegraphics[width=.49\textwidth]{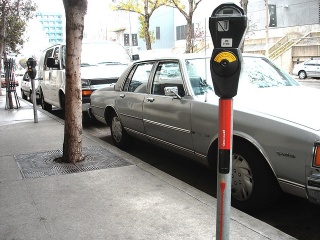} 
    \includegraphics[width=.49\textwidth]{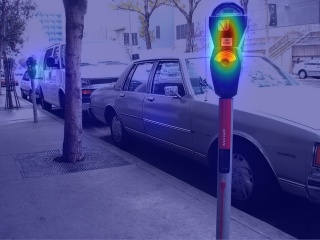}  
    \scriptsize{
        \textbf{Ground-truth}: A car is parked next to a parking meter. \\
        \textbf{Saliency+Context Attention}: A car is parked in the street next to a parking meter. \\
        \textbf{Without saliency}: A car parked next to a white fire hydrant. \\
    }
    \end{minipage}
    \hspace{0.02cm}
    \begin{minipage}[t]{0.32\linewidth}
    \includegraphics[width=.49\textwidth]{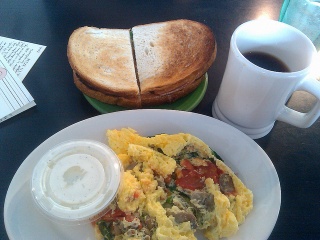} 
    \includegraphics[width=.49\textwidth]{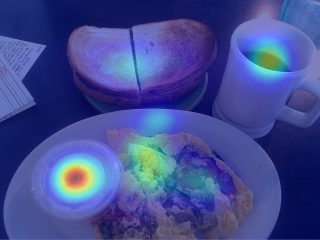}  
    \scriptsize{
        \textbf{Ground-truth}: A plate of food and a cup of coffee. \\
        \textbf{Saliency+Context Attention}: A plate of food with a sandwich and a cup of coffee. \\
        \textbf{Without saliency}: A table with a variety of food on it. \\
    }
    \end{minipage}
\end{center}
\caption{Example results on the Microsoft COCO dataset~\cite{lin2014microsoft}.}
\label{fig:samples}
\end{figure}

\subsection{Qualitative results}
Finally, in Figure~\ref{fig:samples} we report some sample results on images taken from the Microsoft COCO dataset. For each image we report the corresponding saliency map, and captions generated by our model and by the Soft Attention baseline compared to the ground-truth. It can be seen that, on average, captions generated by our model are more consistent with the corresponding image and the human-generated caption, and that, as also observed in the previous section, salient parts are described as well as the context. The incorporation of saliency and context also help the model to avoid failures due to hallucination, such as in the case of the fourth image, in which the Soft Attention model predicts a remote control which is not depicted in the image. Other failure cases, which are avoided by our model, include the repetition of words (as in the fifth image) and the failure to describe the context (first image). We speculate that the presence of two separate attention paths, which the model has learned to attend during the generation of the caption, helps to avoid such failures more effectively than the classic machine attention approach. 

\begin{figure}[t]
\begin{center}
    \begin{minipage}[t]{0.32\linewidth}
    \includegraphics[width=.49\textwidth]{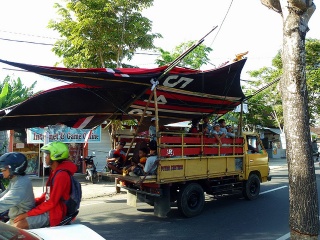}  
    \includegraphics[width=.49\textwidth]{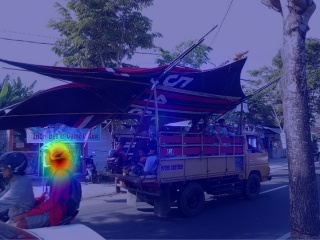}  
    \scriptsize{
        \textbf{Ground-truth}: The yellow truck passes by two people on motorcycles from opposing directions. \\
        \textbf{Saliency+Context Attention}: A person on a motor bike in a city. \\
        \textbf{Without saliency}: A man in a red shirt on a horse. \\
    }
    \end{minipage}
    \hspace{0.025cm}
    \begin{minipage}[t]{0.32\linewidth}
    \includegraphics[width=.49\textwidth]{images/failures/COCO_val2014_000000262509.jpg}  
    \includegraphics[width=.49\textwidth]{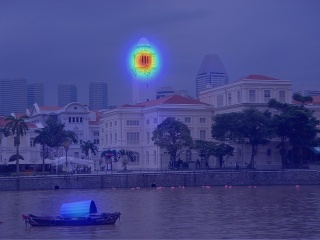}  
    \scriptsize{
        \textbf{Ground-truth}: A cityscape that is seen from the other side of the river. \\
        \textbf{Saliency+Context Attention}: A large building with a large clock tower in the background. \\
        \textbf{Without saliency}: A large building with a large clock in the water. \\
    }
    \end{minipage}
    \hspace{0.025cm}
    \begin{minipage}[t]{0.32\linewidth}
    \includegraphics[width=.49\textwidth]{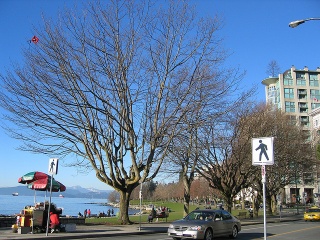} 
    \includegraphics[width=.49\textwidth]{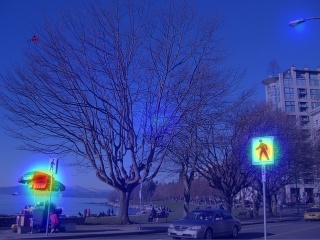}  
    \scriptsize{
        \textbf{Ground-truth}: A large tree situated next to a large body of water. \\
        \textbf{Saliency+Context Attention}: A person is sitting under a red umbrella. \\
        \textbf{Without saliency}: A street sign with a large tree in the middle. \\
    }
    \end{minipage}
    \begin{minipage}[t]{0.32\linewidth}
    \includegraphics[width=.49\textwidth]{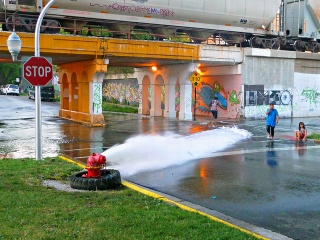}  
    \includegraphics[width=.49\textwidth]{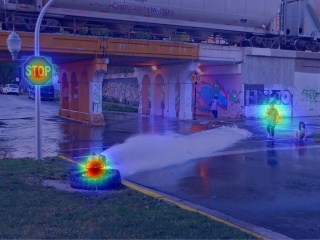}  
    \scriptsize{
        \textbf{Ground-truth}: A busted fire hydrant spewing water out onto a street. \\
        \textbf{Saliency+Context Attention}: A person standing in a front of a large cruise ship. \\
        \textbf{Without saliency}: A man is standing in a dock near a large truck. \\
    }
    \end{minipage}
    \hspace{0.025cm}
    \begin{minipage}[t]{0.32\linewidth}
    \includegraphics[width=.49\textwidth]{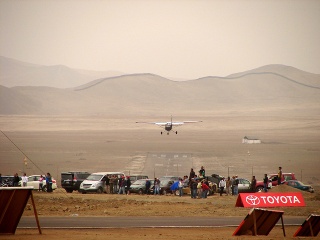}  
    \includegraphics[width=.49\textwidth]{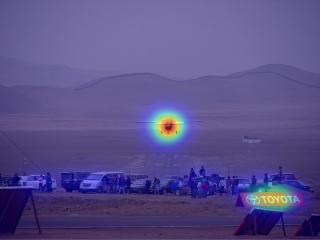}  
    \scriptsize{
        \textbf{Ground-truth}: A small airplane flying over a field filled with people. \\
        \textbf{Saliency+Context Attention}: A group of people walking around a large jet. \\
        \textbf{Without saliency}: A large group of people standing on top of a lush green field. \\
    }
    \end{minipage}
    \hspace{0.025cm}
    \begin{minipage}[t]{0.32\linewidth}
    \includegraphics[width=.49\textwidth]{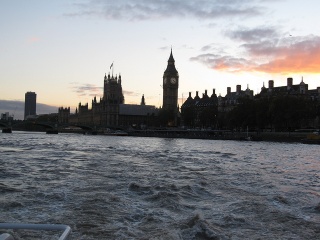} 
    \includegraphics[width=.49\textwidth]{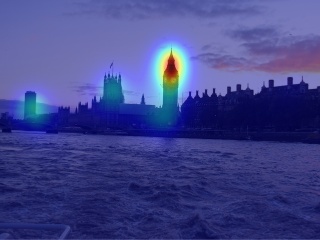}  
    \scriptsize{
        \textbf{Ground-truth}: The view of city buildings is seen from the river. \\
        \textbf{Saliency+Context Attention}: A large clock tower towering over the water. \\
        \textbf{Without saliency}: A large building with a large clock tower in the water. \\
    }
    \end{minipage}
\end{center}
\caption{Failure cases on sample images of the Microsoft COCO dataset~\cite{lin2014microsoft}.}
\label{fig:failures}
\end{figure}

For completeness, some failure cases of the proposed model are reported in Figure~\ref{fig:failures}. The majority of failures occurs when the salient regions of the image are not described in the corresponding ground-truth caption (as for example in the first row), thus causing a performance loss. Some problems arise also in presence of complex scenes (such as in the fourth image). However, we observe that the Soft Attention baseline fails as well to predict correct and complete captions in these cases.  

\section{Conclusion}
We proposed a novel image captioning architecture which extends the machine attention paradigm by creating two attentive paths conditioned on the output of a saliency prediction model. The first one is focused on salient regions, and the second on contextual regions: the overall model exploits the two paths during the generation of the caption, by giving more importance to salient or contextual regions as needed. The role of saliency with respect to context has been investigated by collecting statistics on semantic segmentation datasets, while the captioning model has been evaluated on large scale captioning datasets, using standard automatic metrics and by evaluating the diversity and the dictionary size of the generated corpora. Finally, the activations of the two attentive paths have been investigated, and we have shown that they correspond, word by word, to a focus on salient objects or on the context in the generated caption; moreover, we qualitatively assessed the superiority of the captions generated by our method with respect to those generated by the Soft Attention approach. Although our focus has been that of demonstrating the effectiveness of saliency on captioning, rather than that of beating captioning approaches which rely on different cues, we point out that our method can be easily incorporated into those architectures.

\bibliographystyle{ACM-Reference-Format}
\bibliography{bibliography}

\end{document}